\documentclass{article}

\usepackage[nonatbib,preprint]{neurips_2020}
\usepackage[utf8]{inputenc} 
\usepackage[T1]{fontenc}    
\usepackage{hyperref}       
\usepackage{url}            
\usepackage{booktabs}       
\usepackage{amsfonts}       
\usepackage{nicefrac}       
\usepackage{microtype}      
\usepackage{graphicx}
\usepackage{subfigure}

\usepackage{bbm}
\usepackage{color}
\usepackage{multirow}
\usepackage{amsmath,amssymb}
\usepackage{caption}
\usepackage{bm}
\usepackage[table ]{ xcolor}
\usepackage{color}
\usepackage{enumitem}
\title{Monocular 3D Detection with Geometric Constraints Embedding and Semi-supervised Training}

\author{Peixuan Li\\
  University of Chinese Academy of Sciences\\
  \texttt{lipeixuan@sia.cn} \\
}

\begin{document}

\maketitle

\begin{abstract}
  In this work, we propose a novel single-shot and keypoints-based framework for monocular 3D objects detection using only RGB images, called KM3D-Net. 2D detection only requires a deep neural network to predict 2D properties of objects, as it is a semanticity-aware task. For image-based 3D detection, we argue that the combination of the deep neural network and geometric constraints are needed to estimate appearance-related and spatial-related information synergistically. Here, we design a fully convolutional model to predict object keypoints, dimension, and orientation, and then combine these estimations with perspective geometry constraints to compute position attribute. Further, we reformulate the geometric constraints as a differentiable version and embed it into the network to reduce running time while maintaining the consistency of model outputs in an end-to-end fashion. Benefiting from this simple structure, we then propose an effective semi-supervised training strategy for the setting where labeled training data is scarce. In this strategy, we enforce a consensus prediction of two shared-weights KM3D-Net for the same unlabeled image under different input augmentation conditions and network regularization. In particular, we unify the coordinate-dependent augmentations as the affine transformation for the differential recovering position of objects and propose a keypoints-dropout module for the network regularization. Our model only requires RGB images without synthetic data, instance segmentation, CAD model, or depth generator. Nevertheless, extensive experiments on the popular KITTI 3D detection dataset indicate that the KM3D-Net surpasses all previous state-of-the-art methods in both efficiency and accuracy by a large margin. And also, to the best of our knowledge, this is the first time that semi-supervised learning is applied in monocular 3D objects detection. We even surpass most of the previous fully supervised methods with only 13\% labeled data on KITTI.
\end{abstract}
\section{Introduction}
This work focuses on 3D object detection using only monocular RGB image for autonomous driving. 3D object detection plays an essential role in serving autonomous vehicle perception and robotic navigation. Most of the existing methods heavily rely on LiDARs \cite{chen2017multi,zhou2018voxelnet,yang2018pixor,qi2018frustum,shi2019pointrcnn} data for obtaining accurate depth information. However, LiDAR systems have some disadvantages such as expensive price, high energy consumption, short working life, and adverse to the shape of current vehicles. Alternatively, every car has an onboard monocular camera, and it is cost-effective, energy-conservation, and installation-flexible. It therefore drawn an increasing attention to computer vision community in recent years \cite{brazil2019m3drpn,chen2016monocular,manhardt2019roi,mousavian20173d,qin2019monogrnet,simonelli2019disentangling,li2020rtm3d,he2019mono3d++}.

3D detection from only one monocular image is a naturally ill-posed problem for the reason that missing the depth information introduces ambiguities of inverse projection from the 2D image plane to 3D space. Straight-forward solutions simultaneously regress object dimension, orientation, and position by designing a convolutional neural network(CNN)\cite{xu2018multi,Li_2019_CVPR,zhou2019objects}. 
One central issue in these approaches
is the limited capacity of deep learning techniques to estimate 3D spatial information of large search fields without the help of depth information. To address these challenges, recent works are attempting to apply a geometric constraint as a post-processing step to help CNNs compute 3D position\cite{Naiden2019ShiftRD,liu2019deep,brazil2019m3drpn,li2019stereo,mousavian20173d,li2020rtm3d}.
This disconnected setup makes the gradient cannot be transmitted back to the CNN during the training phase, leading the absence in the perception of geometric constraints. And also, their weak constraints from 3D corners to the 2D edge (exhaustively enumerate $8^4 = 4096$ configurations to determine one result) need accurate but time-consuming two-stage 2D detectors. Besides, the nonlinear or discrete optimization widely adopted in post-processing modules further increases the running time, which largely determines the safety of autonomous driving.

In this paper, we present a single-stage and end-to-end convolutional neural network for accurate and efficient 3D object detection using only monocular RGB images, named KM3D-Net. Our formulation combines the strengths of both CNNs and perspective geometry constraints in one unified framework. To pave the way for this, we split 3D properties into appearance- and spatial-related information. The former includes keypoint, dimension, and orientation, which can be obtained easily from image space by exploiting semantic information or visual change.  Keypoints are defined as the perspective projection from the center and corners of the 3D bounding box(BBox). It can provide 18 point-to-point constraints, much stronger than the four uncertainty constraints in the 2D BBox.
The appearance-related information is predicted by adding extra parallel branches after feature extraction in an excellent 2D detector \cite{zhou2019objects}. For spatial-related information, we reformulate the non-linear optimization in the projection space as a task of solving the overdetermined system, which can pass error differentials back to CNN via using the singular value decomposition(SVD) operator.
Importantly, with our reformulation, the whole framework, which comprises a CNN and geometric reasoning, can maintain the consistency between appearance- and spatial-related prediction in an end-to-end fashion during the training phase. Our method extends the ability of CNN by explicitly embedding perspective geometric model rather than increasing the depth or width, making it possible to improve accuracy while maintaining high efficiency.

Semi-supervised training has been proven to be effective by image classification in exploiting unlabeled data\cite{Berthelot2019MixMatch,laine2016temporal,NIPS2015_5947,tarvainen2017mean}. However, it received little attention in 3D detection where labeling data is much more expensive. To bridge this gap, we propose a semi-supervised training method for the scene where 3D annotations are scarce by extending our simple KM3D-Net. Our method is inspired by $\Pi$-model\cite{laine2016temporal}, a semi-supervised baseline method of image classification. It proposes that different predictions of a good model should be assembled at the same input with different network regularization and input augmentation. In this paper, we show how to extend this idea to the 3D detection task. In particular, we propose a Keypoints-Dropout that randomly drops the keypoints in our geometric reasoning module for the network regularization instead of regular Dropout \cite{srivastava2014dropout}. The rationalization of this dropping is that one keypoint provides two constraints and at least two keypoints will be able to compute the position information. Our Keypoints-Dropout can be regarded as an explicit dropout of a known feature. For input augmentation, we formulate the coordinate-dependent augmentation, such as rotation, translation, flipping, or scaling, as the affine transformation to align the coordinates of one object in two different augmentations. These two strategies allow our model to predict a stochastic variable in the same input. Then our unsupervised loss penalizes the difference of 3D properties provided by our KM3D-Net using the mean square difference. The final semi-supervised loss combines supervised and unsupervised loss by a time-dependent weighting function.

Our contribution is summarized as follows: 1) A simple and efficient architecture combines the strengths of both CNN and perspective geometry and also achieves real-time 3D objects detection using only monocular images. 2) A differentiable geometric reasoning module can be embedded in CNN for the estimation of spatial-related information while maintaining the consistency of CNN output. 3) A semi-supervised training method exploits unlabeled images in monocular 3D objects detection. 4) Experiments on the popular KITTI 3D detection benchmark demonstrate that the proposed method outperforms the state-of-the-art competitors by large margins in both accuracy and speed.
\section{Related Work}
Existing approaches could be divided into two groups: 1) detecting methods that include category-specific 3D shape prior, extra data, or stand-alone network. 2) detecting methods that only using RGB images. \\
\textbf{Monocular 3D Detection Including Extra Data.} Mono3D \cite{chen2016monocular} focuses on generating 3D object proposals by using instance segmentation, object contour, and ground-plane assumption. 
To remedy the lack of scene depth, alternative methods \cite{xu2018multi, ma2019accurate, manhardt2019roi, he2019mono3d++,wang2019pseudo} incorporate stand-alone depth to extend image information. Given the depth map, these methods detect 3D object either by joint optimization \cite{he2019mono3d++}, or by multi-stage fusion\cite{xu2018multi,ma2019accurate,manhardt2019roi}, or by LiDAR-based methods after transforming it into point clouds \cite{wang2019pseudo}. Another family methods represent cars as the keypoints in the form of a wire-frame template and use external annotated CAD models to produce synthetic data for training. \cite{chabot2017deep, zeeshan2014cars,murthy2017reconstructing, he2019mono3d++}. Although extra network and annotated data would increase the accuracy of the detection, they also need labor-intensive annotations work and additional computation in both training and inference time. Such computation may not be appropriate for unmanned vehicle and robots with limited computational resources.\\
\textbf{Monocular 3D Detection Using RGB Images Only.}
Compared with the aforementioned approaches, most recent works try to detect 3D objects from only RGB images. Deep3DBox \cite{mousavian20173d} infers 2D BBox, dimension, and orientation, and then generates 3D position by combining geometric constraints of 3D points to 2D lines. GS3D \cite{Li_2019_CVPR} predicts the guidance of cuboid by adding extra 3D information branch in Faster-RCNN \cite{ren2015faster}, and then extract feature from the projected region of guidance to perform accurate refinement. MonoGRNet \cite{qin2019monogrnet} optimizes the 3D BBox in an end-to-end manner by fusion of 2D detection, instance depth estimation, 3D location estimation, and local corner regression.
M3D-PRN \cite{brazil2019m3drpn} proposes a 3D proposals network with a depth-aware convolution to generate 2D and 3D object proposals simultaneously, and use 3D-2D geometric constraints as a post-processing module to improve precision.
MonoDIS \cite{simonelli2019disentangling} isolate the group parameters to simplifies the training dynamics by disentangling the 3D transformation. All these methods incorporate a 2D proposal network as the base architecture, which cause the bottlenecks of inference time.
The most related approach to ours is RTM3D \cite{li2020rtm3d} that proposes to use keypoints as the intermediate and combines the geometric to further refine the estimation inferred by a single-stage network. Although the fast CNN structure is adopted, its geometric constraint module is still time-consuming and non-differentiable which prevents it from jointly optimizing the output of the CNN.\\
\textbf{Semi-supervised Training.} Semi-supervised training has promising power in image classification for improving model performance when large labeled datasets are not available \cite{reddy2018semi-supervised,tu2019a}. However, these methods have not been extended to the scope of monocular 3D detection, whose labeling cost is much more expensive. Popular approaches in semi-supervised training are based on consistency-regularization \cite{lee2013pseudo,rasmus2015semi,laine2016temporal,sajjadi2016mutual,tarvainen2017mean,miyato2018virtual,Berthelot2019MixMatch,xie2019unsupervised,xie2019self,sohn2020fixmatch}, which enforce the model to generate the same output when its inputs and model are perturbed. For image classification, the key idea is to manually combine different image transformations for input augmentation and employ dropout \cite{srivastava2014dropout} in fully connected layers for network regularization. Extending these methods to coordinate-aware 3D detection tasks, especially efficient 3D detection, face two main differences: 1) 3D detection requires to align the properties of one 3D objects in two coordinates-dependent augmentations for the reason that these augmentations change the 2D position and camera intrinsic matric. 2) Efficient 3D detection with a fully convolutional network requires another paradigm to replace dropout that only works well in the fully connected network.

\section{Monocular 3D Objects Detection}
\begin{figure}
  \centering
  \includegraphics[width=0.95\columnwidth]{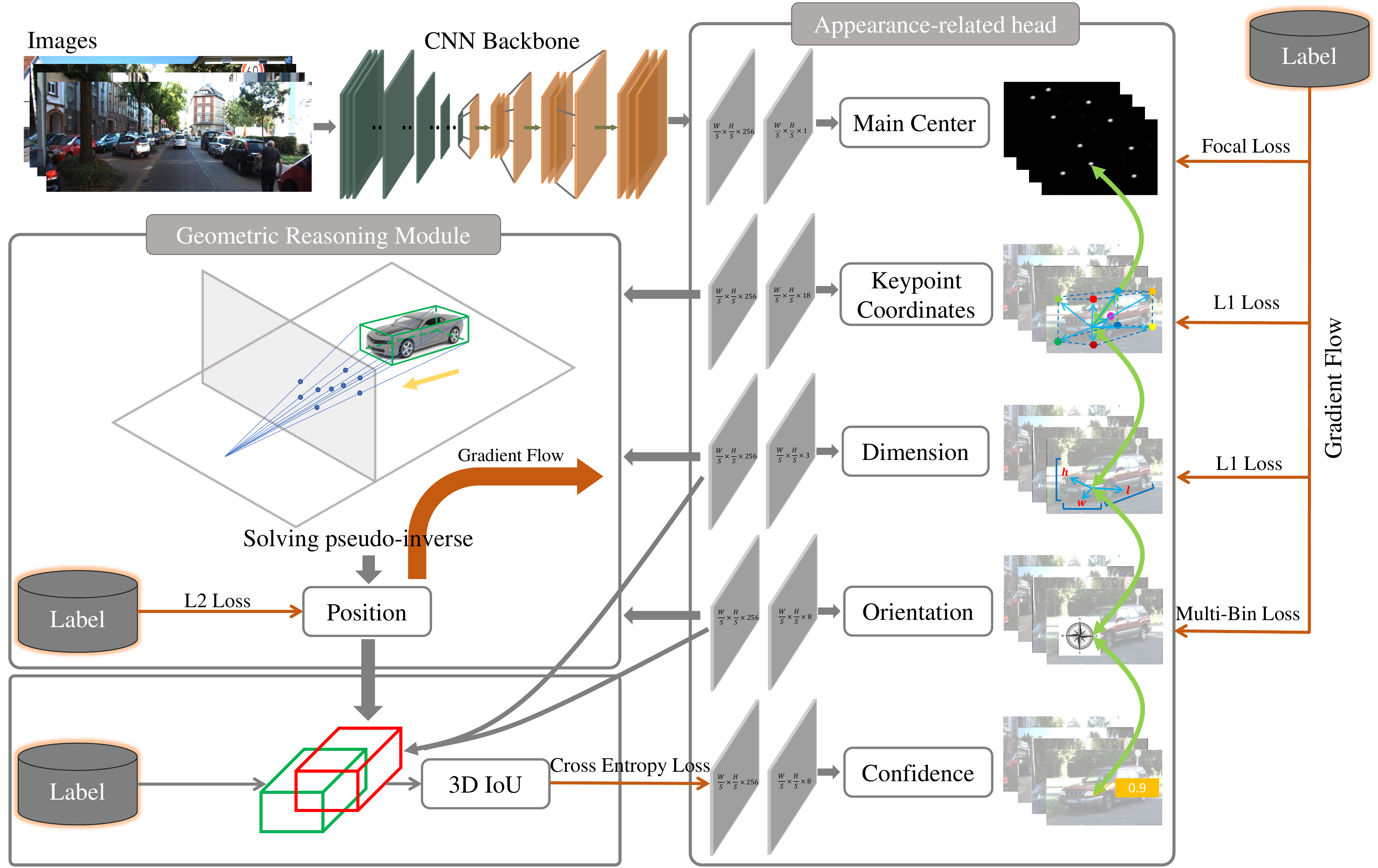}
  \caption{Overview of proposed KM3D-Net which output keypoints, object dimensions, local orientation, and 3D confidence, followed by our differentiable geometric consistency constraints to predict position.}
  \label{fig:supervision}
\end{figure}
Our method comprises a fully convolutional network stage, which predicts appearance-related properties of an object: dimension, orientation, and ordered list of 2D perspective keypoints, followed by a geometric reasoning module, which
performs the point-to-point geometric constraints for 3D position prediction. The complete framework combines the strengths of both CNNs and perspective geometry and is trainable in end-to-end fashion via the standard back-propagation algorithm \cite{Lecun98gradient-basedlearning}.
As shown in Fig. \ref{fig:supervision}, our model contains a backbone architecture to extract the feature of the entire image and a series of detection head:
\begin{itemize}
\setlength{\itemsep}{0pt}
\setlength{\parsep}{0pt}
\setlength{\parskip}{0pt}
\item \textbf{Main center}: We associate an object and its properties by a single point at 2D BBox center, called the main center. The head produce a main center heatmap $M \in R^{\frac{h}{4} \times \frac{w}{4} \times c}$ where $c$ is the object categories.
\item \textbf{Keypoints}: Keypoints detection head estimates 9 ordered 2D perspective keypoints projected from the center and corners of 3D object BBox. These keypoints are geometrically and semantically consistent across different instances of certain object categories. We estimate these keypoints by the regression of offset coordinate $M_k \in R^{\frac{h}{4} \times \frac{w}{4} \times 18}$ from the main center.  
\item \textbf{Dimension}: It regresses the residual value $\delta_D \in R^{\frac{h}{4} \times \frac{w}{4} \times 3}$ to restore object dimension by $\bar{H}e^{\delta_H}, \bar{W}e^{\delta_W}, \bar{L}e^{\delta_L}$ followed \cite{simonelli2019disentangling}, where $\bar{H}=1.63, \bar{W}=1.53, \bar{L}=3.88$ are statistical average size in the KITTI dataset.
\item \textbf{Orientation}: It regresses the local orientation $O_l \in R^{\frac{h}{4} \times \frac{w}{4} \times 8}$ with respect to the ray through the perspective point of 3D center instead of global orientation followed \emph{Multi-Bin} based method \cite{mousavian20173d}.
\item \textbf{3D Confidence}: It predict the 3D BBox confidence $P_{3D} \in R^{\frac{h}{4} \times \frac{w}{4} \times 1}$ by the IoU between estimation and ground truth.
The quality of the final 3D BBox is also related to the confidence of the main center. We combine these in the Bayesian rule $Pro=Pro_{2D}^{m}*Pro_{3D}$ to obtain the final 3D confidence. $Pro_{2D}^{m}$ are extracted by the corresponding heatmap after the sigmoid function.
\end{itemize}
During the training step, we define the multi-task loss as:
\begin{equation}
	\label{eq:a}
    L_{sup}=\omega_{m}L_{m}+\omega_{kc}L_{kc}+\omega_{D}L_{D}+\omega_{O}L_{O}+\omega_{T}L_{T}+\omega_{conf}L_{conf}
	\end{equation}
where the main center heatmap loss $L_m$ are the penalty reduced focal loss followed \cite{zhou2019objects,law2018cornernet,li2020rtm3d}. Keypoints offset coordinate loss $L_{kc}$ is a depth-guided $L1$ loss for dynamically adjusting punish coefficient of different scaling objects. See the supplementary material for more details. Dimension loss $L_D$ is an ordinary $L1$ loss with respect to the ground truth. Orientation loss $L_D$ is the \emph{Multi-Bin}  \cite{mousavian20173d,zhou2019objects}, which split the whole orientation field to two bin and employ a hybrid discrete-continuous loss for the training of each bin. 3D confidence loss $L_{conf}$ is self-supervised by the IoU between the predicted 3D BBox and ground truth with the standard binary cross-entropy loss.

Given the predicted 9 keypoints $\widehat{kp}$, dimension $\widehat{D}$ and orientation $\widehat{O}$, the object position $T$ can be solved by minimizing the re-projection error of keypoints from 3D BBox corners and center.
\begin{equation}
	\label{eq:a}
\widehat{T}=\mathop{\arg\min}\limits_{T}\sum\limits_{i}^{9}\left\|kp^{'}_i(T,\widehat{D},\widehat{O})-\widehat{kp_i}\right\|_2
	\end{equation}
where $kp^{'}(T,\widehat{D},\widehat{O})$ can be defined as:
\begin{equation}
	\label{eq:solvingposition}
    \begin{aligned}
	\renewcommand*{\arraystretch}{1.5}
    {kp^{'}(T,\widehat{D},\widehat{O})}=K
    \left[
    \begin{matrix}
    R(\widehat{O})^{3 \times 3}& T^{1 \times 3}\\
    0^T&1
    \end{matrix}
    \right]diag(\widehat{D})Cor\\
     Cor=\left[\begin{smallmatrix}
    0   & 0   & 0    &  0 &-1  &  -1 &  -1 &  -1 & -1/2  \\
    1/2 & -1/2& -1/2 &1/2 &1/2 &-1/2 &-1/2 &1/2  &  0   \\
    1/2 & 1/2 & -1/2 &-1/2&1/2 &1/2  &-1/2 &-1/2 &  0\\
    1   & 1   &1     & 1  &1   &1    &1    &1    &  1
    \end{smallmatrix}\right]
    \end{aligned}
	\end{equation}
In order to embed this geometry relationship into our KM3D-Net, we first normalize the $\widehat{kp}$ to $\widetilde{kp}$ by camera intrinsic for simplifying representation. Then re-arrange Eq. \ref{eq:solvingposition} to $R(\widehat{O})diag(\widehat{D})Cor+T-\widetilde{kp}=0$ and reformulate it into a linear system :
\begin{equation}
  \begin{aligned}
  \left[
  \begin{matrix}
  -1& 0& \widetilde{kp}_1^x\\
   0&-1& \widetilde{kp}_1^y\\
   &\vdots&\\
   -1& 0& \widetilde{kp}_9^x\\
   0&-1& \widetilde{kp}_9^y\\
  \end{matrix}\right]_{18\times3}\left[
  \begin{matrix}
  X\\
  Y\\
  Z\\
  \end{matrix}\right]=\left[
  \begin{matrix}
	\frac{l}{2}\cos(\theta)+\frac{w}{2}\sin(\theta)-\widetilde{kp}_1^x(-\frac{l}{2}\sin(\theta)+\frac{w}{2}\cos(\theta))\\
    \frac{h}{2}-\widetilde{kp}_1^y(-\frac{l}{2}\sin{\theta}+\frac{w}{2}\cos(\theta))\\
   \vdots\\
   0\\
    0\\
   \end{matrix}\right]_{18 \times 1}
    \end{aligned}	
\end{equation}
Supplementary material contains the complete equation. The position $T=[X, Y, Z]^T$ can be recovered through pseudo inverse using the SVD operator. We backpropagate it by using matrix calculus\cite{Giles2008An,Ionescu2015Matrix}.
Equipping this differentiable geometric reasoning module, the position error $L_{T}=\left\|\widehat{T}-T \right\|_2$ can be passed through keypoints, dimension and orientation to maintain the inherent consistency between them.

\section{Semi-supervised approach}
\begin{figure}
  \centering
  \includegraphics[width=1\columnwidth]{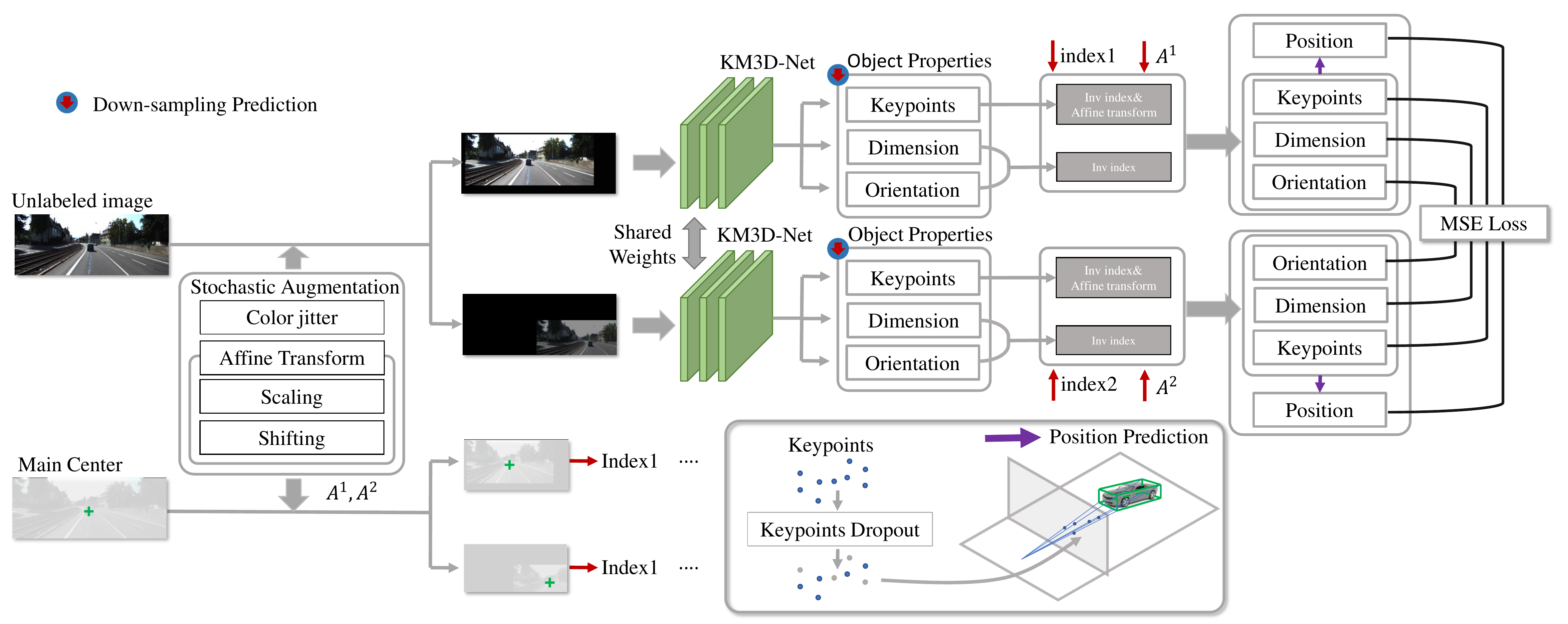}
  \caption{Overview of our unsupervised training. It leverages affine transformation to unify input augmentation and devise keypoints dropout for regularization. These two strategies make proposed KM3D-Net output two stochastic variables with the same input. Penalizing their differences is our training goal.}
  \label{fig:semi-supervision}
\end{figure}
We propose a semi-supervised training approach to further improve the model accuracy for the scene where labeled training data is scarce. We train our model in a semi-supervised fashion by containing labeled and unlabeled data in a batch to optimize the model jointly.

Unlabeled data training is shown in Fig. \ref{fig:semi-supervision}. We use KM3D-Net to evaluate the same input image twice with different augmentation and regularization. The unsupervised loss penalizes different estimation for the same object in the image by taking the mean square difference(MSE loss). Input augmentation consists of two components: coordinate-independent and coordinate-dependent. The first component is randomly color jittering. The second component includes randomly horizontal flip, shifting, and scaling. We formulate these operations as the affine transformation to convert or restore the coordinate of the main center and keypoints. This uniform expression of the coordinate-dependent augmentation as a matrix form can satisfy the network's differentiability. In addition, we propose keypoints dropout for network regularization. It randomly drops keypoints in the processing of position prediction. It is important to notice that, one keypoint can provide two geometric constraints, and at least two keypoints can compute the position information of three degrees of freedom. Thus dropping some of the 9 keypoints is reasonable in computing position information. Proposed keypoints dropout can be regarded as a special case of dropout \cite{srivastava2014dropout} and has two benefits: 1) make the model more accurate in predicting the un-dropout keypoints 2) the inference that contains all the keypoints has more generalization power.
The input augmentation and keypoints dropout make the same network weights output a random variable in the training step. Given the same input, their difference can be seen as an optimization goal. Mathematical, we define the unsupervised loss as:
  \begin{equation}
  \label{eq:a}
    L_{unsup}=\sum_{i \in L\cup U} \sum_{P,R,D} MSE\left( A_1^{-1}f\left(Augment\left(I_i; A_1 \right); \theta \right)-A_{2}^{-1}f\left(Augment\left(I_i; A_2\right); \theta \right)\right)
  \end{equation}
  where $L$ and $U$ are labeled set and unlabeled set respectively. $A_1$ and $A_2$ are two different affine transformation of augmentation. $f(\theta)$ is the trainable KM3D-Net. Then semi-supervised loss is computed by $L_{ssl}= L_{sup}+ \omega(t) L_{unsup}$, where $\omega(t)$ is Gaussian ramp-up curve $ \exp[-5( 1-t/ 100)^2]$ following\cite{laine2016temporal}.
  \section{Experiments}
  \subsection{Dataset and Implementation Details}
  We evaluate our experiments on the KITTI 3D detection benchmark \cite{geiger2012we}, which contains 7481 labeled training images and 7518 unlabeled test images. Since the ground truth of the test set is not available, we train and evaluate our model in three splitting following in \cite{mousavian20173d,Li_2019_CVPR,qin2019monogrnet,liu2019deep,brazil2019m3drpn}: $train1, val1$;  $train2, val2$ and $train, test$.
  We experiment with 2 backbone for the trade-off between speed and accuracy: ResNet-18 \cite{he2016deep} and DLA-34 \cite{yu2018deep}.
  We implemented our deep neural network in PyTorch and trained by using the Adam optimizer with a base learning rate of $0.0001$ for $200$ epochs and reduce $10\times$ at $100$ and $180$ epochs. We then project the 3D BBox of ground truths in the coordinate frame of the left and right image and use random scaling (between 0.6 to 1.4), random shifting in the image range, and color jittering as data augmentation. In the inference step, after imposing a $3\times3$ max pooling on the heatmap of the main center and keypoints, we first extract the object location by filtering the main center with threshold 0.4 and then using it to select the remaining parameters from other output.
  \subsection{Results}
\begin{figure}
		\begin{center}
			\includegraphics[width=1\columnwidth]{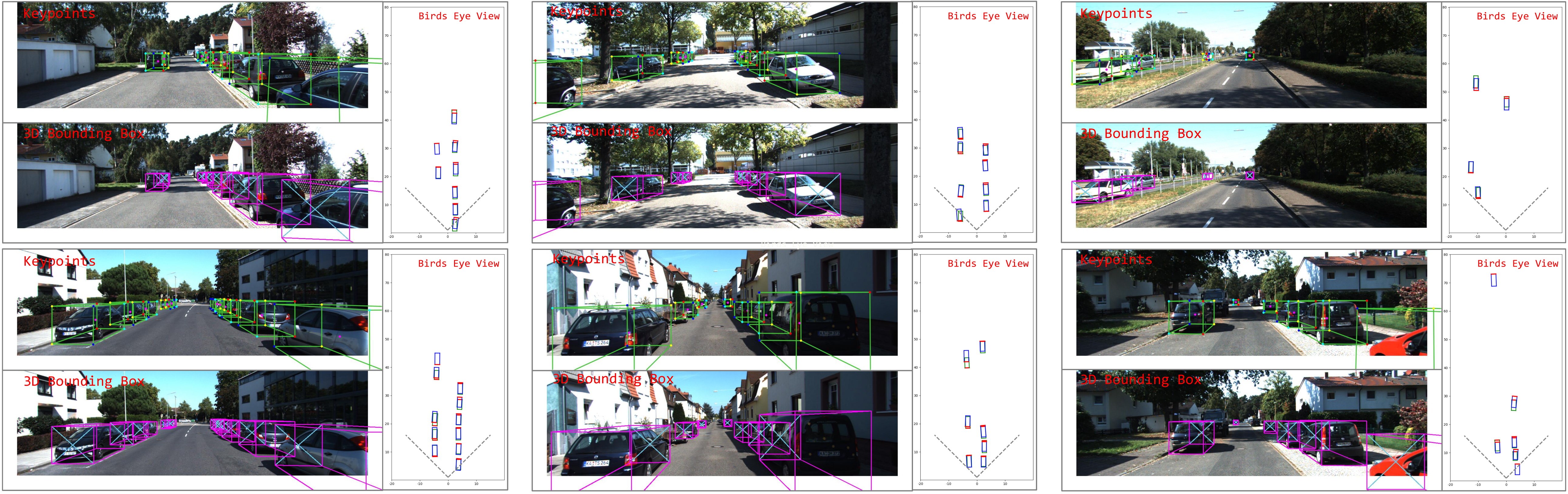}
		\end{center}
     \caption{\textbf{
     Qualitative results.} In birds eye view, blue box is predicted value, and green box is the ground truth.}
		\label{fig:qualitative}
  \end{figure}

  We report three official evaluation metrics on the KITTI dataset: average precision for 3D IoU ($AP_{3D}$), Birds Eye View ($AP_{BEV}$), and 2D IoU ($AP_{2D}$, See the supplementary material for its results.). Since KITTI has used  $AP^{40}$ instead of $AP^{11}$ and existing approaches only report accuracy at 11 points on the $val_1$ or $val_2$ set, we thus give results of the average precision under 40 points in the test set and 11 points in $val1$ and $val2$ set for a fair comparison.
  \begin{table*}[htb]
    \tiny
    \caption{Comparison 3D detection methods for car category, evaluated by metric $AP_{3D}$ on \bm{$val_1$} / \bm{$val_2$} / \bm{$test$} set on KITTI. “Extra” means the extra training data. The best, second, and third results are highlighted in \textcolor{red}{red}, \textcolor{blue}{blue} and \textcolor{cyan}{cyan} respectively.}
    \begin{center}
    \resizebox{\textwidth}{!}{
    \begin{tabular}{| c  | c  | c | c@{/}c | c@{/}c | c@{/}c | c@{/}c@{/}c | c@{/}c@{/}c | c@{/}c@{/}c |}
    \hline
    \multirow{2}{*}{Method} & \multirow{2}{*}{Extra} & \multirow{2}{*}{Time} & \multicolumn{6}{c|}{$IoU>0.5$     [$ \bm{val_1/val_2$}]} & \multicolumn{9}{c|}{$IoU>0.7$ [\bm{$val_1/val_2/$\colorbox{yellow}{$test$}}]} \\
    \cline{4-18}
    & & & \multicolumn{2}{c|}{Easy} & \multicolumn{2}{c|}{Moderate} & \multicolumn{2}{c|}{Hard} & \multicolumn{3}{c|}{Easy} & \multicolumn{3}{c|}{Moderate} & \multicolumn{3}{c|}{Hard} \\
    \hline
  Mono3D \cite{chen2016monocular}(CVPR2016) & Mask & 4.2 s & 25.19 &-& 18.20 &- & 15.52 & - & 2.53 & -&- & 2.31 &-& - & 2.31&-&- \\
  3DOP \cite{chen20173d}(TPAM2017) & Stereo & 3 s &   46.04&- & 34.63 & - &  30.09 & - & 6.55 & -& - & 5.07 & -& - & 4.10 &-& -\\
  MF3D \cite{xu2018multi}(CVPR2018) & Depth & - & 47.88 & 45.57 & 29.48 & 30.03 & 26.44 & 23.95 & 10.53 & 7.85 & 7.08& 5.69 & 5.39 & 5.18 & 5.39 & 4.73 & 4.68 \\
  Mono3D++ \cite{he2019mono3d++}(AAAI2019) & Depth+Shape & $>$0.6s & 42.00 & - & 29.80 & - & 24.20 & - & 10.60 & - &- & 7.90 & - & - &5.70 & - &- \\
  ROI-10D \cite{manhardt2019roi}(CVPR2019) & Depth & - & 37.59 & - & 25.14 & - & 21.83 & - & 9.61 & - &12.30 & 6.63 & - & 10.30 &6.29 & - &9.39 \\
  \hline
  \hline
  Deep3DBox \cite{mousavian20173d}(CVPR2017) & None & - & 27.04 & - & 20.55 &  -& 15.88 & - & 5.85&- & - & 4.10& - & - & 3.84 & - & -\\
  GS3D \cite{Li_2019_CVPR}(CVPR2019) & None &2.3s& 32.15 & 30.60 & 29.89 & 26.40& 26.19 & 22.89 & 13.46 & 11.63& 4.47& 10.97 & 10.51 & 2.90& 10.38  & 10.51 & 2.47 \\
  MonoGRNet \cite{qin2019monogrnet}(AAAI2019) & None & \textbf{\textcolor{cyan}{0.06s}} & \textbf{\textcolor{blue}{50.51}} & - & \textbf{\textcolor{cyan}{36.97}} & - & 30.82 & - & 13.88 & -& 9.61 & 10.19 & -& 5.74 & 7.62 & -& 4.25 \\
  FQNet\cite{liu2019deep}(CVPR2019)& None & 3.33s & 28.16 & 28.98 & 21.02 & 20.71 & 19.91 & 18.59 & 5.98 & 5.45& 2.77 & 5.50 & 5.11& 1.51 & 4.75 & 4.45 & 1.01\\
  ROI-10D \cite{manhardt2019roi}(CVPR2019) & None & - & 29.38 & - & 19.80 & - & 18.04 & - & 10.12 & - &- & 1.76 & - & - &1.30 & - &- \\
  MonoDIS \cite{simonelli2019disentangling}(CVPR2019) & None &-&  {{-}} & {{-}} &  {{-}} & {{-}} & {{-}} & -& {{18.05}} & {{-}}& 10.37& {{14.96}} & {{-}} & 7.96& {{13.42}} & {{-}} & 6.40\\
  M3D-RPN \cite{brazil2019m3drpn}(ICCV2019) & None &0.16s&  \textbf{\textcolor{cyan}{48.96}} & \textbf{\textcolor{blue}{49.89}} & \textbf{\textcolor{blue}{39.57}} & \textbf{\textcolor{blue}{36.14}} & \textbf{\textcolor{blue}{33.01}} & \textbf{\textcolor{blue}{28.98}} & \textbf{\textcolor{blue}{20.27}} & \textbf{\textcolor{blue}{20.40}}& \textbf{\textcolor{blue}{14.76}}& \textbf{\textcolor{blue}{17.06}} & \textbf{\textcolor{blue}{16.48}} & \textbf{\textcolor{blue}{9.71}}& \textbf{\textcolor{blue}{15.21}} & \textbf{\textcolor{blue}{13.34}} & \textbf{\textcolor{blue}{7.42}}\\
  \hline
  Ours(ResNet18) & None & \textbf{\textcolor{red}{0.021s}} & 47.23   & \textbf{\textcolor{cyan}{47.13}}  & 34.12  & \textbf{\textcolor{cyan}{33.31}} & \textbf{\textcolor{cyan}{31.51}}  & \textbf{\textcolor{cyan}{25.84}} & \textbf{\textcolor{cyan}{19.48}} & \textbf{\textcolor{cyan}{18.34}} & \textbf{\textcolor{cyan}{12.65}}& \textbf{\textcolor{cyan}{15.32}} &
  \textbf{\textcolor{cyan}{14.91}}& \textbf{\textcolor{cyan}{8.39}} &
  \textbf{\textcolor{cyan}{13.88}}& \textbf{\textcolor{cyan}{12.58}} &
  \textbf{\textcolor{cyan}{7.12}} \\
  Ours(DLA34)& None & \textbf{\textcolor{blue}{0.040s}} & \textbf{\textcolor{red}{56.02}}  & \textbf{\textcolor{red}{54.09}} & \textbf{\textcolor{red}{43.13}} & \textbf{\textcolor{red}{43.07}}& \textbf{\textcolor{red}{36.77}} & \textbf{\textcolor{red}{37.56}} & \textbf{\textcolor{red}{22.50}} & \textbf{\textcolor{red}{22.71}} &
  \textbf{\textcolor{red}{16.73}}& \textbf{\textcolor{red}{19.60}} &
  \textbf{\textcolor{red}{17.71}}& \textbf{\textcolor{red}{11.45}} &
  \textbf{\textcolor{red}{17.12}}& \textbf{\textcolor{red}{16.15}} &
  \textbf{\textcolor{red}{9.92}} \\
  \hline
 \end{tabular}}
 \label{tab:3d}
 \end{center}
 \end{table*}
 \begin{table*}[htb]
  \caption{ Comparison of $AP_{BEV}$ for car category.}
  \begin{center}
  \resizebox{\textwidth}{!}{
  \begin{tabular}{| c | c | c | c@{/}c | c@{/}c | c@{/}c | c@{/}c@{/}c | c@{/}c@{/}c | c@{/}c @{/}c|}
  \hline
  \multirow{2}{*}{Method} & \multirow{2}{*}{Accelerator} & \multirow{2}{*}{FPS} & \multicolumn{6}{c|}{$IoU>0.5$     [$ \bm{val_1/val_2$}]} & \multicolumn{9}{c|}{$IoU>0.7$ [\bm{$val_1/val_2/$\colorbox{yellow}{$test$}}]}\\
  \cline{4-18}
  & & & \multicolumn{2}{c|}{Easy} & \multicolumn{2}{c|}{Moderate} & \multicolumn{2}{c|}{Hard} & \multicolumn{3}{c|}{Easy} & \multicolumn{3}{c|}{Moderate} & \multicolumn{3}{c|}{Hard} \\
  \hline
  Mono3D \cite{chen2016monocular}(CVPR2016) & - & 0.2 & 30.50 & - & 22.39& - & 19.16 & -  & 5.22 & -& - & 5.19 &-& - & 4.13 & - & -\\
  3DOP \cite{chen20173d}(TPAM2017)& Titan X & 0.3&  55.04 & - &  41.25 & - &  34.55 & - & 12.63 & -& - & 9.49 & -& - & 7.59 & -& - \\
  MF3D \cite{xu2018multi}(CVPR2018) &  Titan X&- & 55.02 & 54.18 & 36.73 & 38.06 & 31.27 & 31.46 & 22.03 & 19.20 &13.73& 13.63 & 12.17 &9.62 & 11.60 & 10.89&8.22 \\
  Mono3D++ \cite{he2019mono3d++}(AAAI2019) & Titan X & $>$1.7 & 46.70 & - & 34.30 & - & 28.10 & - & 16.70 & - & -&11.50 & -&- & 10.10 & -&- \\
  ROI-10D \cite{manhardt2019roi}(CVPR2019) & - & - & 46.85 & - & 34.05 & - & 30.46 & - & 14.50 & - &16.77 & 9.91 & - & 12.40 &8.73 & - &11.39 \\
  \hline
  \hline
  Deep3DBox \cite{mousavian20173d}(CVPR2017) & - & - & 30.02 & - & 23.77 & - & 18.83 & -& 9.99 & -& - & 7.71& -&- & 5.30 & - &-\\
  GS3D \cite{Li_2019_CVPR}(CVPR2019) & - &0.4& - &- &- &- &- &- &- &- &8.41&- &-& 6.08&- &-&4.94  \\
  MonoGRNet \cite{qin2019monogrnet}(AAAI2019) &Tesla P40 & \textbf{\textcolor{cyan}{16.7}} & - &- &- &- &- &- &- &-& 18.19&- &-&11.17 &- &-&8.73 \\
  FQNet\cite{liu2019deep}(CVPR2019)& 1080Ti & 0.3 & 32.57 & 33.37 & 24.60 & 26.29 & 21.25 & 21.57& 9.50 & 10.45 &5.40& 8.02 & 8.59 &3.23& 7.71 & 7.43&2.46 \\
  ROI-10D \cite{manhardt2019roi}(CVPR2019) & - & - & 36.21 & - & 24.90 & - & 21.03 & - & 14.04 & - &- & 3.69 & - & - &3.56 & - &- \\	
  MonoDIS \cite{simonelli2019disentangling}(CVPR2019) & V100 &-&  {{-}} & {{-}} &  {{-}} & {{-}} & {{-}} & -& {{24.26}} & {{-}}& 18.45& {{18.43}} & {{-}} & 12.58& {{16.95}} & {{-}} & 10.66\\
  M3D-RPN \cite{brazil2019m3drpn}(ICCV2019) & 1080Ti &6.3& \textbf{\textcolor{blue}{55.37}} & \textbf{\textcolor{blue}{55.87}} & \textbf{\textcolor{blue}{42.49}} & \textbf{\textcolor{blue}{41.36}} & \textbf{\textcolor{blue}{35.29}} & \textbf{\textcolor{blue}{34.08}} & \textbf{\textcolor{blue}{25.94}} & \textbf{\textcolor{blue}{26.86}}&\textbf{\textcolor{blue}{21.02}} & \textbf{\textcolor{blue}{21.18}} & \textbf{\textcolor{blue}{21.15}} &\textbf{\textcolor{blue}{13.67}}& \textbf{\textcolor{blue}{17.90}} & \textbf{\textcolor{blue}{17.14}}& \textbf{\textcolor{cyan}{10.23}}\\
  \hline
  Ours(ResNet18) & 1080Ti & \textbf{\textcolor{red}{47.6}} & \textbf{\textcolor{cyan}{53.77}}  & \textbf{\textcolor{cyan}{51.92}} & \textbf{\textcolor{cyan}{40.58}}  & \textbf{\textcolor{cyan}{39.69}} & \textbf{\textcolor{cyan}{34.79}} & \textbf{\textcolor{cyan}{34.07}} & \textbf{\textcolor{cyan}{24.48}} & \textbf{\textcolor{cyan}{24.86}} & \textbf{\textcolor{cyan}{19.71}} & \textbf{\textcolor{cyan}{19.10}} & \textbf{\textcolor{cyan}{17.44}} & \textbf{\textcolor{cyan}{13.37}}&\textbf{\textcolor{cyan}{16.54}}&\textbf{\textcolor{cyan}{16.83}}&\textbf{\textcolor{blue}{11.10}} \\
  Ours (DLA34)& 1080Ti & \textbf{\textcolor{blue}{25.0}} & \textbf{\textcolor{red}{62.39}}  & \textbf{\textcolor{red}{59.35}} & \textbf{\textcolor{red}{49.93}}  & \textbf{\textcolor{red}{45.14}} & \textbf{\textcolor{red}{43.73}} & \textbf{\textcolor{red}{42.47}} & \textbf{\textcolor{red}{27.83}}& \textbf{\textcolor{red}{28.87}} &\textbf{\textcolor{red}{23.44}}& \textbf{\textcolor{red}{23.38}} & \textbf{\textcolor{red}{22.87}} &\textbf{\textcolor{red}{16.20}}& \textbf{\textcolor{red}{21.69}} &
  \textbf{\textcolor{red}{22.55}}&\textbf{\textcolor{red}{14.47}} \\
  \hline
  \end{tabular}}

  \label{tab:BEV}
  \end{center}
  \end{table*}\\
  \textbf{3D Object Detection with Supervised Training.} Table \ref{tab:3d} and \ref{tab:BEV} show the published results for $AP_{3D}$ and $AP_{BEV}$ with fully supervised training on KITTI data set. We also report their single image inferencing time and accelerator. We principally focus on the car class as most previous studies do, and multi-class results can be seen in the supplementary material. Our method outperforms all previous methods in terms of $AP_{3D}$ and $AP_{BEV}$ on both accuracy and running speed. Specifically, DLA-34 as the backbone achieves the best accuracy and is faster than all previous methods. Using ResNet-18, we achieve the best speed with 47 FPS while accuracy also outperforms most of the approaches. The improvement of run time comes from the discarding of region proposal process scheme and the embedding of geometric consistency constraint in CNN. Better accuracy indicates that combining the respective strengths of neural networks and geometry can achieve better performance than using them alone.\\
\textbf{Semi-Supervised Approach.}
For the semi-supervised training, we use $train1$ set and testing set to be the labeled and unlabeled images, and then validate the performance on $val1$ set. Since we are the first semi-supervised method for monocular 3D detection, we mainly compare our semi-supervised training with supervised training for various experimental settings. Fig. \ref{tab:semi-supervised} shows that our semi-supervised technique produces a more efficient performance as the number of labeled data decreases. Note that with our semi-supervised training only 500 labeled data can achieve better accuracy than most previous methods that adopt supervised training with complete annotation set.
\begin{table*}[htb]
  \tiny
  \caption{Semi-supervised training with different amounts of labeled data.}
  \begin{center}
  \resizebox{\textwidth}{!}{
  \begin{tabular}{| c | c@{/}c@{/}c | c@{/}c@{/}c | c@{/}c@{/}c | c@{/}c@{/}c | c@{/}c@{/}c | c@{/}c@{/}c |}
  \hline
  &\multicolumn{18}{c|}{$AP_{3D}^{11}$ IoU>0.5 (Easy/Moderate/Hard)}\\
  \hline
  \textbf{Labeled data}&\multicolumn{3}{c|}{\textbf{100}}&\multicolumn{3}{c|}{\textbf{300}}&\multicolumn{3}{c|}{\textbf{500}}&\multicolumn{3}{c|}{\textbf{1000}}&\multicolumn{3}{c|}{\textbf{2000}}&\multicolumn{3}{c|}{\textbf{3518}}\\
  \hline
  Supervised-only&11.49&8.68&6.48&15.01&11.05&10.35 &22.10&16.09&14.13 &38.97&28.76&24.75 &40.46&29.07&24.71 &47.23&34.12&31.51\\
  \hline
  Semi-supervised with keypoints dropout&12.34&8.99&7.07& 16.50&11.87&10.88 &25.16&18.36&15.96 &39.25&29.11&25.85 &41.15&29.44&25.38 &47.78&34.80&31.32\\
  \hline
  Semi-supervised with augmentation&15.01&10.58&9.23& 21.34&15.66&13.81 &31.73&22.72&18.12 &41.59&30.16&26.76 &44.91&32.90&26.51 &48.09&35.24&31.96\\
  \hline
  Semi-supervised with augmentation and keypoints dropout&16.47&11.12&10.79&22.75 &16.31&15.25 &32.35&23.26&21.19 &41.51&30.31&26.10 &45.66&32.69&27.88 &48.78&36.35&31.68\\
  \hline
  \end{tabular}}

\label{tab:semi-supervised}
\end{center}
\end{table*} \\
\textbf{Qualitative Results.}
The qualitative results of 3D object detection, keypoints estimation,
and bird’s eye view are shown in Fig. \ref{fig:qualitative}. Our method performs accurate 3D detection even for extremely crowded and truncated objects.
\subsection{Ablation Study}
\begin{figure}
\tiny
\begin{minipage}{0.45\columnwidth}
  \centering
      \includegraphics[width=1\columnwidth]{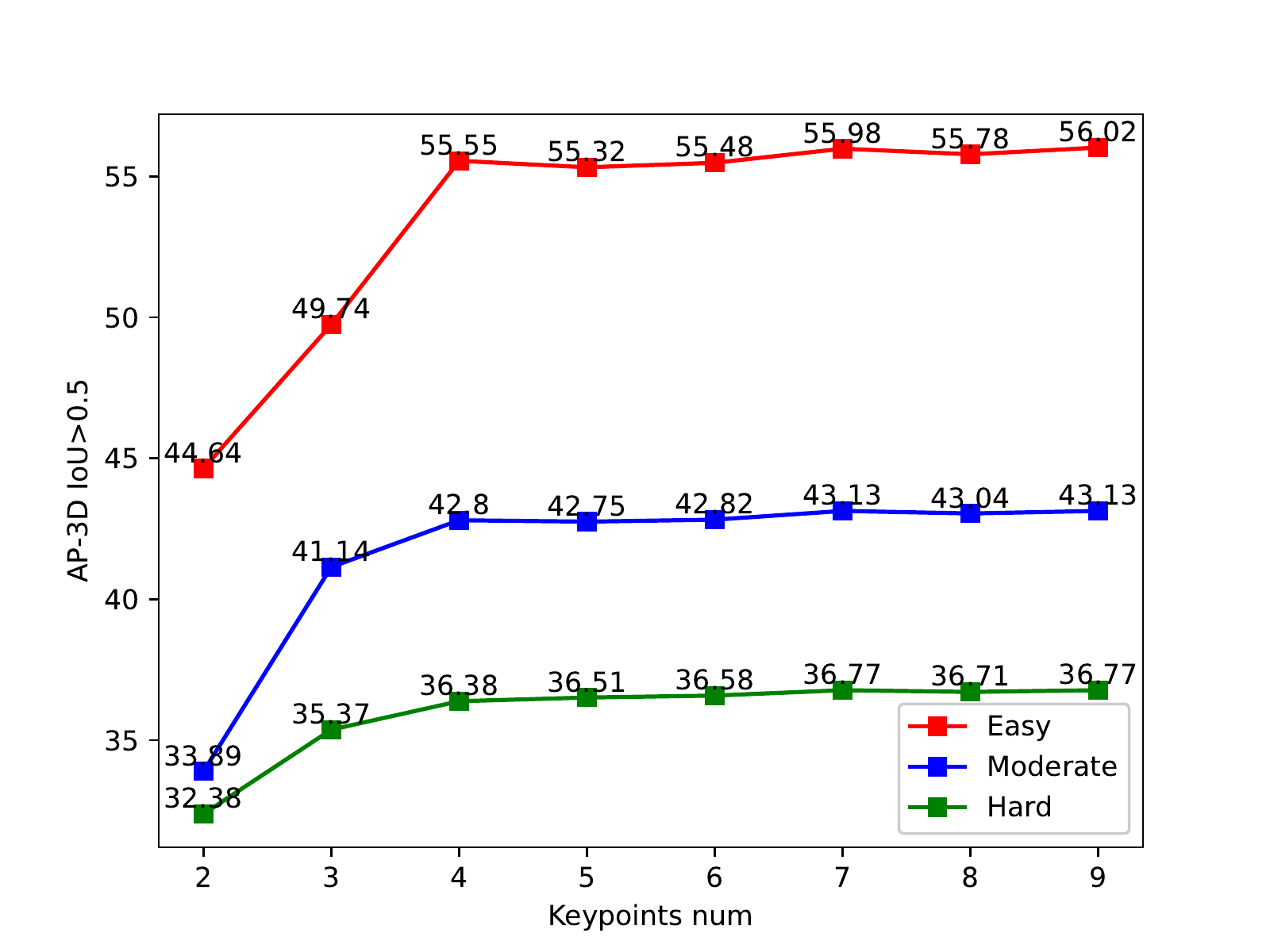}
     \caption{\textbf{Extreme testing.} Input different num of regressed keypoints to our geometric reasoning module.}
    \label{fig:robust}
\end{minipage}\begin{minipage}{0.55\columnwidth}
\centering
\begin{tabular}{|c|c|cc|cc|}
  \hline
        Config & Set & {AP$_{\rm bev}^{0.5}$} &{AP$_{\rm bev}^{0.7}$} & {AP$_{\rm 3d}^{0.5}$} & {AP$_{\rm 3d}^{0.7}$}\\
        \hline
        \multirow{3}{*}{w/o GRM in training} & Easy & 47.47 & 21.77 & 42.26 & 17.33 \\
        \, &  Mode & 35.17 & 17.18 & 32.20 & 14.11 \\
        \, &  Hard & 32.96 & 16.46 & 27.34 & 12.67 \\
        \hline
        \multirow{3}{*}{w/o GRM in all phase} & Easy & 32.70 & 11.28 & 21.38 & 6.94 \\
        \, &  Mode & 28.88 & 10.95 & 19.59 & 5.82 \\
        \, &  Hard & 26.99 & 9.77 & 17.85 & 4.46 \\
        \hline
        \multirow{3}{*}{Prediction in heatmap} & Easy & 52.01 & 23.94 & 47.27 & 15.93 \\
        \, &  Mode & 36.63 & 17.55 & 33.89 & 12.36 \\
        \, &  Hard & 33.52 & 16.28 & 27.95 & 10.92 \\
        \hline
        \multirow{3}{*}{w/o 3D confidence} & Easy & {53.11} & {23.39} & {45.76} & {17.88} \\
        \, &  Mode & {39.42} & {17.37} & {33.24} & {14.38} \\
        \, &  Hard & {33.99} & {16.38} & {30.92} & {12.61} \\
        \hline
        \multirow{3}{*}{w/o Augmentation} & Easy & {45.23} & {18.31} & {40.78} & {12.58} \\
        \, &  Mode &{33.10} &{14.94} & {28.48} & {14.94} \\
        \, &  Hard & {28.05} &{12.78} &{25.55} & {9.87} \\
        \hline
    \end{tabular}
    \captionof{table}{Ablation study of using different strategey.}
    \label{tab:Ablation}
\end{minipage}
\end{figure}

We conduct a series of experiments to validate the contribution of different strategies in our model: \textbf{A1}) Geometry reasoning module (GRM). Two settings will be considered to verify GRM's validity. One is only to use GRM in the inference phase and remove GRM of position supervision in the training phase. Another is to remove GRM in all phases. In this case, we add a parallel depth prediction branch after feature extraction and combine the main center to compute position;
\textbf{A2}) The keypoint coordinates of regression vs. heatmap. For prediction keypoints in the heatmap, we add a parallel heatmap branch, whose each channel contains the corresponding semantic points of all objects in the whole picture. We provide more details in the
supplementary material; \textbf{A3}) 3D confidence prediction; \textbf{A4}) Data augmentation. Table \ref{tab:Ablation} shows the $AP_{3D}^{11}$ and $AP_{BEV}^{11}$ for each variant of the proposed strategies. The performance drops a lot without our geometric reasoning module,
indicating the validity of geometric constraints in spatial-related information prediction. Using heatmaps to estimate keypoints also suffer from performance degradation, especially for the moderate and hard group. We believe that this is because the heatmap prediction of different keypoints are semantically ambiguous, and cannot estimate the keypoints of the truncated region. We provide its visualization in the supplementary material.
\\
\textbf{Extreme testing.} To further illustrate the potential performance of the proposed method, we conducted some extreme testing. We input a different number of keypoints to geometric reasoning module for computing position information. Each keypoints are chosen randomly. As shown in Fig. \ref{fig:robust}, even with only two keypoints, our geometry reasoning module still has an acceptable performance. Such a powerful ability is the main reason for the overall effectiveness of our method.
\section{Conclusion and Future Work}
In this work, we present a novel single-shot framework for monocular 3D object detection. Our formulation embed a differentiable module of perspective geometry in CNN to help promote the running efficiency and optimize outputs of network jointly. Combining the strengths of both CNN and geometric constraints, our method outperform all existing image-only methods in both efficiency and accuracy by large margins. Another contribution of this work is a semi-supervised training method in the settings where labeled data is scarce. Our method only requires unlabeled data and its intrinsic camera parameters, making it practical in scenarios and provide a flexibility for further developing semi-supervised training of 3D object detection.

In the future, we are interested to extend our method to stereo 3D detection and 3D multi-object tracking with semi-supervised training.
\medskip

\clearpage
\appendix
\section{Depth-guided L1 Loss}
Keypoints coordinates of distant and near objects have different error tolerance to compute 3D position. In particular, due to the perspective projection, the keypoints coordinates of near objects have larger value items in regular L1 loss then distant objects. To balance their contributions, we design a depth-guided L1 loss function.
\begin{equation}
  \label{eq:a}
  \begin{aligned}
    L_{kc}=\frac{1}{N} \sum\limits  \sum\limits_{j} \sum\limits_{i}^9 g(Z_j)\left\|kp_j^i-\widehat{kp}_j^i\right\|_1\\
    g(Z)=
\begin{cases}
  \alpha Z   & if \quad Z<a\\
\log_{10}(Z+1-a)+ \alpha a  & if \quad Z\ge a
\end{cases}
  \end{aligned}
  \end{equation}
where we set $\alpha=0.01$ and $a=5$ in our experiments.
\section{Complete Equation in Position Solving}
\begin{equation}
  \begin{aligned}
  \left[s
  \begin{matrix}
  -1& 0& kp_1^x\\
   0&-1& kp_1^y\\
   -1& 0& kp_2^x\\
   0&-1& kp_2^y\\
   -1& 0& kp_3^x\\
   0&-1& kp_3^y\\
   -1& 0& kp_4^x\\
   0&-1& kp_4^y\\
   -1& 0& kp_5^x\\
   0&-1& kp_5^y\\
   -1& 0& kp_6^x\\
   0&-1& kp_6^y\\
   -1& 0& kp_7^x\\
   0&-1& kp_7^y\\
   -1& 0& kp_8^x\\
   0&-1& kp_8^y\\
  \end{matrix}\right]\left[
  \begin{matrix}
  X\\
  Y\\
  Z\\
  \end{matrix}\right]=\left[
  \begin{matrix}
	\frac{l}{2}\cos(\theta)+\frac{w}{2}\sin(\theta)-kp_1^x(-\frac{l}{2}\sin(\theta)+\frac{w}{2}\cos(\theta))\\
    \frac{h}{2}-kp_1^y(-\frac{l}{2}\sin{\theta}+\frac{w}{2}\cos(\theta))\\
    \frac{l}{2}\cos(\theta)-\frac{w}{2}\sin(\theta)-kp_2^x(-\frac{l}{2}\sin(\theta)-\frac{w}{2}\cos(\theta))\\
    \frac{h}{2}-kp_2^y(-\frac{l}{2}\sin{\theta}-\frac{w}{2}\cos(\theta))\\
    -\frac{l}{2}\cos(\theta)-\frac{w}{2}\sin(\theta)-kp_3^x(\frac{l}{2}\sin(\theta)-\frac{w}{2}\cos(\theta))\\
    \frac{h}{2}-kp_3^y(\frac{l}{2}\sin{\theta}-\frac{w}{2}\cos(\theta))\\
    -\frac{l}{2}\cos(\theta)+\frac{w}{2}\sin(\theta)-kp_4^x(\frac{l}{2}\sin(\theta)+\frac{w}{2}\cos(\theta))\\
    \frac{h}{2}-kp_4^y(\frac{l}{2}\sin{\theta}+\frac{w}{2}\cos(\theta))\\
    \frac{l}{2}\cos(\theta)+\frac{w}{2}\sin(\theta)-kp_5^x(-\frac{l}{2}\sin(\theta)+\frac{w}{2}\cos(\theta))\\
    -\frac{h}{2}-kp_5^y(-\frac{l}{2}\sin{\theta}+\frac{w}{2}\cos(\theta))\\
    \frac{l}{2}\cos(\theta)-\frac{w}{2}\sin(\theta)-kp_6^x(-\frac{l}{2}\sin(\theta)-\frac{w}{2}\cos(\theta))\\
    -\frac{h}{2}-kp_6^y(-\frac{l}{2}\sin{\theta}-\frac{w}{2}\cos(\theta))\\
    -\frac{l}{2}\cos(\theta)-\frac{w}{2}\sin(\theta)-kp_7^x(\frac{l}{2}\sin(\theta)-\frac{w}{2}\cos(\theta))\\
    -\frac{h}{2}-kp_7^y(\frac{l}{2}\sin{\theta}-\frac{w}{2}\cos(\theta))\\
    -\frac{l}{2}\cos(\theta)+\frac{w}{2}\sin(\theta)-kp_8^x(\frac{l}{2}\sin(\theta)+\frac{w}{2}\cos(\theta))\\
    -\frac{h}{2}-kp_8^y(\frac{l}{2}\sin{\theta}+\frac{w}{2}\cos(\theta))\\
   \end{matrix}\right]
    \end{aligned}	
\end{equation}
\section{2D Objects Detection}
\begin{figure}[htb]
  \begin{center}
    \includegraphics[width=0.9\columnwidth]{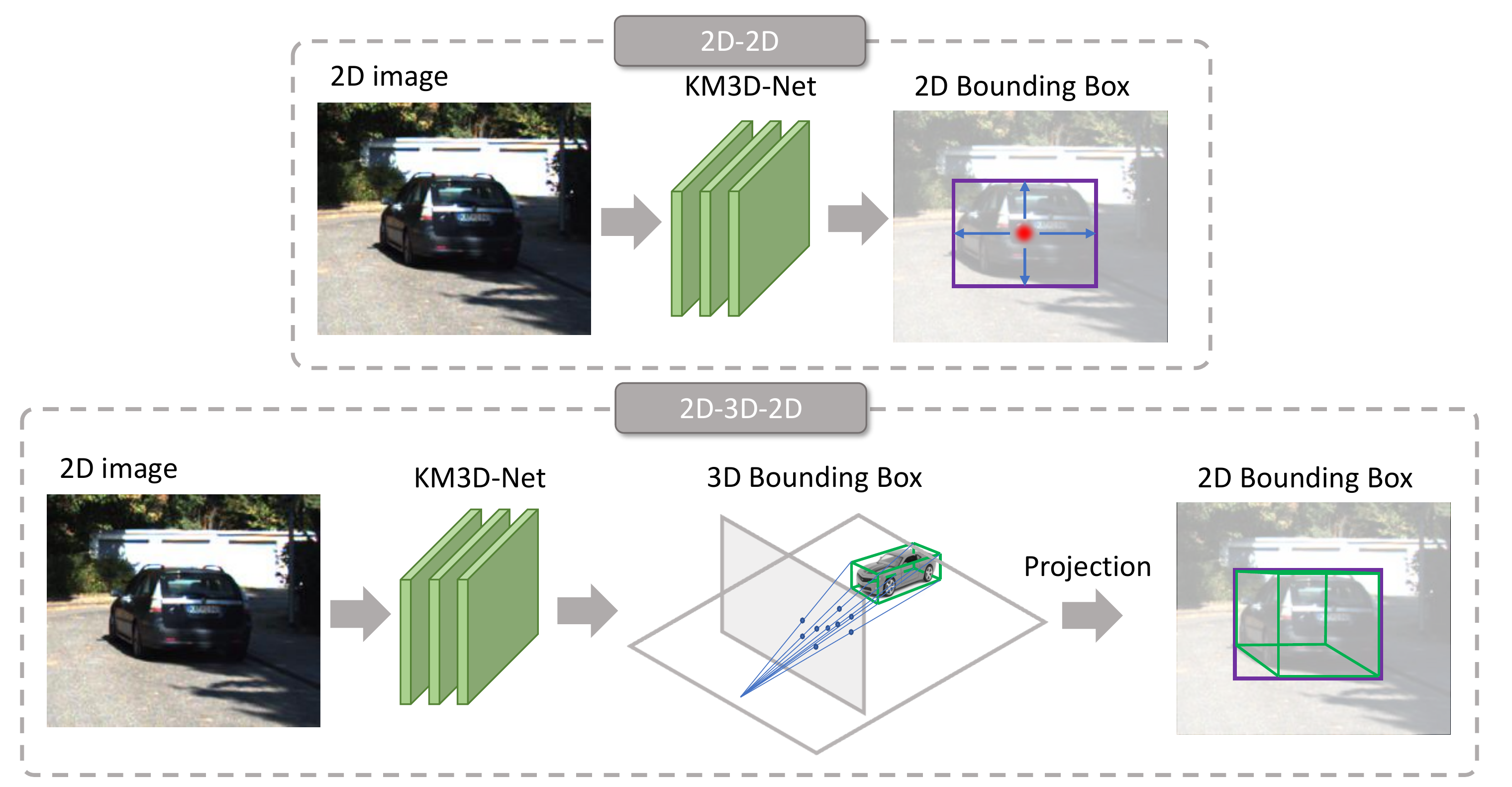}
  \end{center}
   \caption{Comparison between 2D-2D and 2D-3D-2D fashion.}
  \label{fig:2D}
\end{figure}
\begin{figure}[htb]
  \begin{center}
    \includegraphics[width=1\columnwidth]{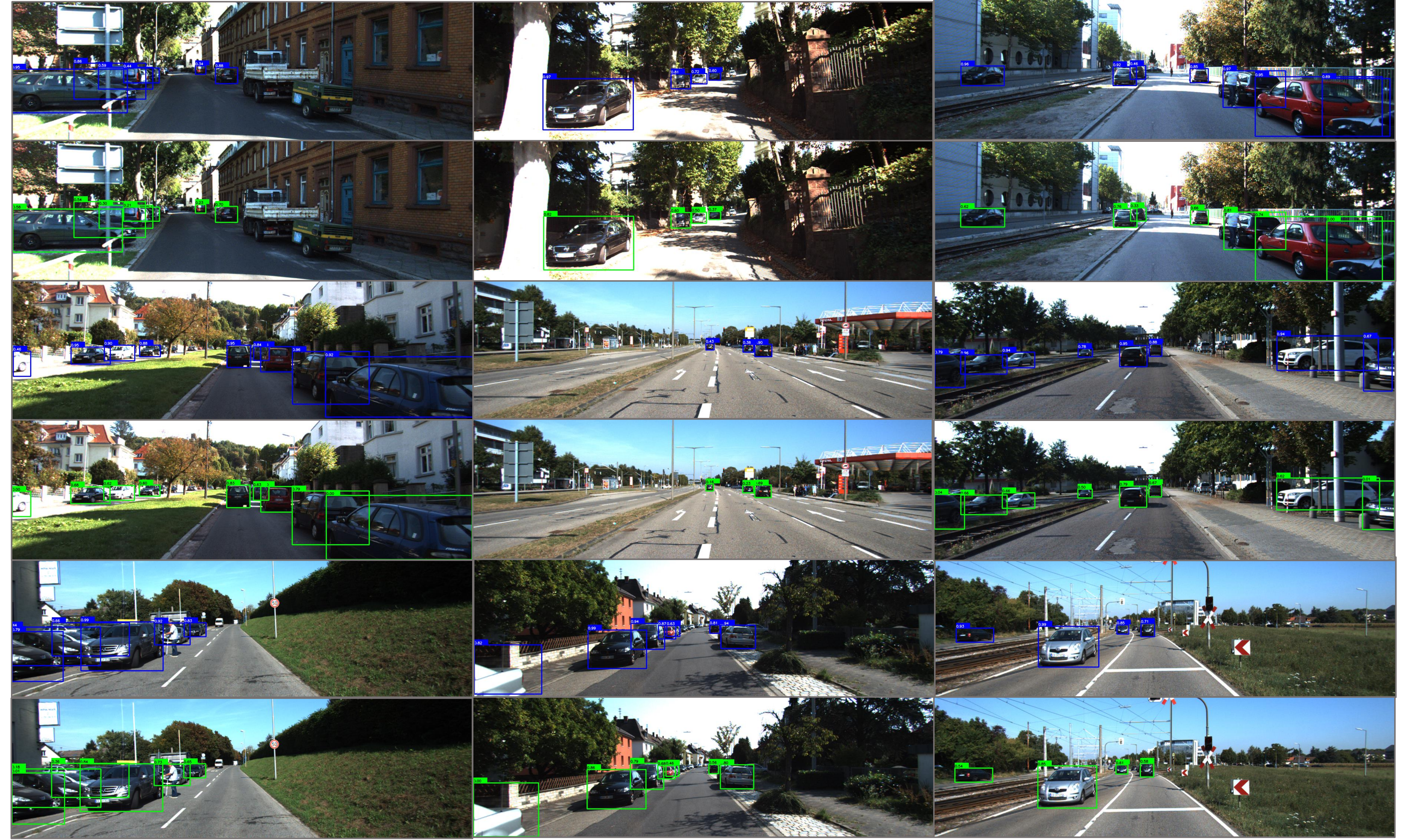}
  \end{center}
   \caption{\textbf{Qualitative results.} The results of 2D-2D fashion are shown in the blue box and 2D-3D-2D fashion in green box. Our new paradigm obtains 2D BBox from 3D space and has a similar result to 2D-2D fashion.}
  \label{fig:2Dvis}
\end{figure}

As shown in Fig. \ref{fig:2D}, there are two ways to generate 2D BBox in our method 1) \textbf{2D-2D:} we add the size prediction head parallel behind the backbone and employ L1 loss for training. 2) \textbf{2D-3D-2D:} We propose a new paradigm to detect 2D object, which obtains 2D BBox from 3D space. Specifically, we first estimate the 3D BBox by KM3D-Net and then obtain 2D BBox by computing the smallest bounding box around the projected corners of 3D BBox on image space. The quantitative and qualitative results are shown in Table \ref{tab:2Ddetection} and Fig \ref{fig:2Dvis}.With the DLA-34 backbone, our 2D-2D fashion outperforms all previous approaches. Our 2D-3D-2D fashion has also achieved a competitive performance, especially for the hard set where the region of truncated and occluded are more serious then easy and moderate set. We believe that this 2D-3D-2D fashion combined with shape and dimension prior has a similar pipeline with human perception. This achievement also highlights the strength of our 3D objects detection.
\begin{table}[htb]
  \scriptsize
  \caption{2D detection $AP_{2D}$ with IoU>=0.7 and orientation AOS results
  for the car category evaluated in KITTI's test set.}
    \centering
    \begin{tabular}{| c |c| c | c c c | c c c |}
      \hline
      \multirow{2}{*}{Method}& \multirow{2}{*}{Extra}&\multirow{2}{*}{Time} & \multicolumn{3}{c|}{$\text{AP}_\text{2D}$} & \multicolumn{3}{c|}{$\text{AOS}$} \\
      \cline{4-9}
      &&&Easy&Moderate&Hard&Easy&Moderate&Hard\\
      \hline
      AM3D \cite{manhardt2019roi}(ICCV2019)&Depth&-&92.55 &	88.71&77.78 &	- &	- &	- \\
      Decoupled-3D \cite{cai2020monocular}(AAAI2019)&Depth&-&87.78 &	67.92 &	54.53 &87.34 &	67.23 &	53.84  \\
      TLNet \cite{qin2019tlnet}(CVPR2019)&Stereo&-&76.92 &	63.53 &	54.58 &&&\\
      MonoFENet \cite{monofenet}(TIP2019)&Depth&0.15&91.68 &	84.63 &	76.71 &	91.42 &	84.09 &	75.93 \\
      MonoPSR \cite{ku2019monopsr}(CVPR2019)&Depth&0.20s&93.63 &88.50 &	73.36
      &	93.29 &87.45 &72.26 \\
      StereoRCNN\cite{licvpr2019}(CVPR2019)&Stereo&-&93.98 &	85.98 &	71.25 &-&-&-\\
ROI-10D \cite{manhardt2019roi}(CVPR2019)&None&-&76.56 &	70.16&61.15 &	75.32 &	68.14 &	58.98 \\
MonoGRNet \cite{qin2019monogrnet}(AAAI2019)&None&0.06s&88.65 &	77.94 &	63.31 &-&-&-\\
FQNet\cite{liu2019deep}(CVPR209)&None&3.33s&94.72 &	90.17 &	76.78 &93.66&87.49&73.61 \\
GS3D \cite{Li_2019_CVPR}(CVPR2019)&None&2.3s& 86.23  & 76.35 & 62.67 & 85.79  & 75.63  & 61.85\\
M3D-RPN \cite{brazil2019m3drpn}(ICCV2019)&None&0.16s&89.04 &	85.08 &	69.26 &	88.38 &	82.81 &	67.08 \\
      \hline
      Ours-ResNet18 (2D-2D) &None&0.021s& 93.35 & 82.97 & 73.11 & 93.13&82.43&72.47 \\
      Ours-ResNet18 (2D-3D-2D) &None&0.021s& 88.59 & 75.78 & 65.85 & 88.53&75.54&65.56\\
      Ours-DLA34 (2D-2D)&None &0.040s& \textbf{96.44} & \textbf{91.07} & \textbf{81.19} & \textbf{96.34}&\textbf{90.70}&\textbf{90.72} \\
      Ours-DLA34 (2D-3D-2D) &None&0.040s& 90.67 & 88.77 & 80.15 & 90.37&88.38&79.35 \\
      \hline
      \end{tabular}
    \label{tab:2Ddetection}
    \end{table}

\section{Multi-class 3D Detection Results}
\begin{figure}[htb]
  \begin{center}
    \includegraphics[width=1\columnwidth]{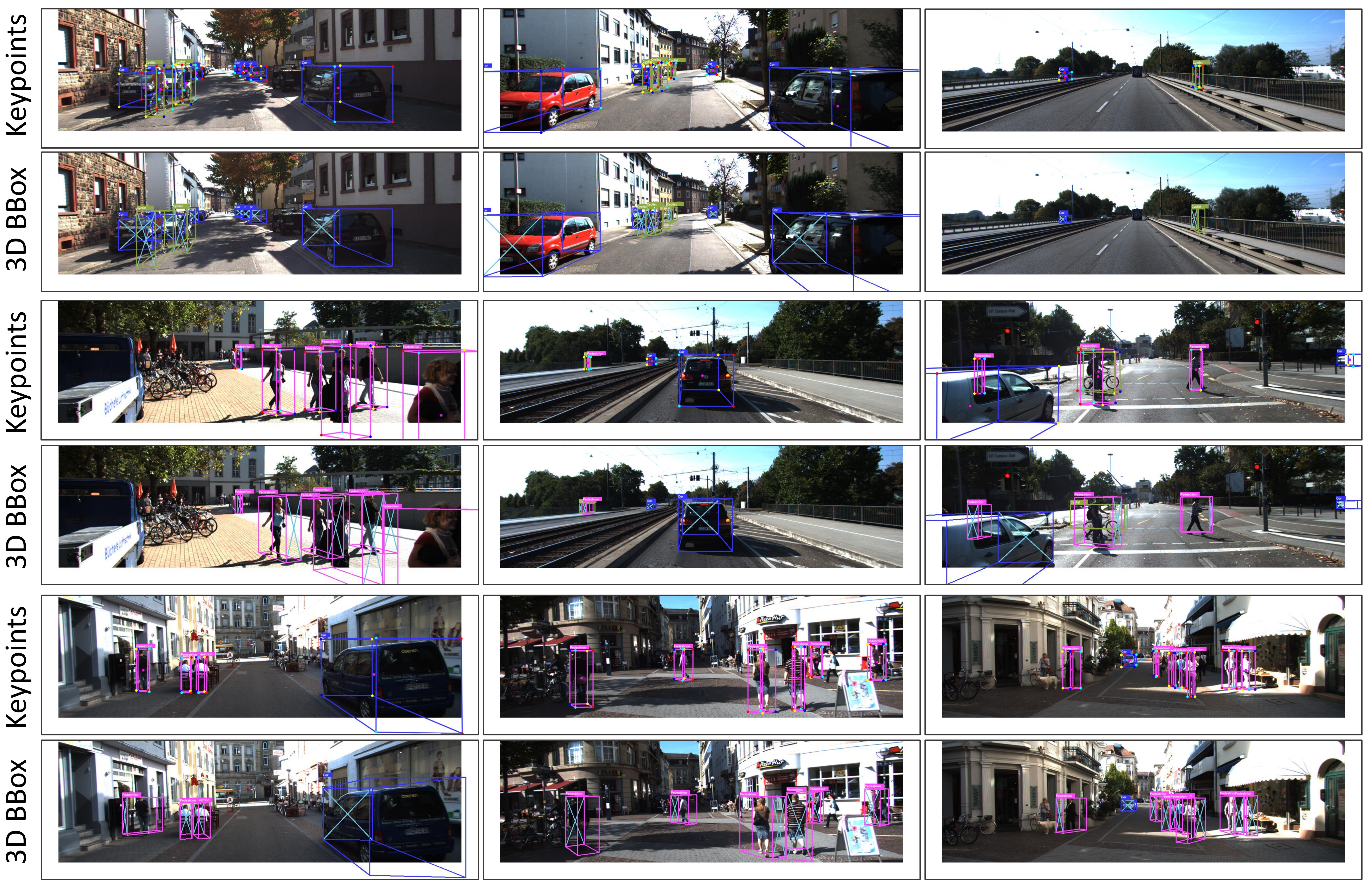}
  \end{center}
   \caption{Qualitative results of multi-class 3D object detection.}
  \label{fig:multi}
\end{figure}

In addition to the car category, we also apply the proposed framework to other categories to demonstrate generalization.
The quantitative and qualitative results with the DLA-34 backbone are shown in Table \ref{tab:2Ddetection} and Fig \ref{fig:multi}.
\begin{table}[htb]
  \tiny
  \caption{Multi-class 3D detection resultis in $val1$ set.}
    \centering
    \begin{tabular}{| c | c c c | c c c |c c c |c c c |}
      \hline
      \multirow{2}{*}{category}& \multicolumn{3}{c|}{$\text{AP}_\text{3D}$ IoU>0.5}  & \multicolumn{3}{c|}{$\text{AP}_\text{BEV}$ IoU>0.5} &\multicolumn{3}{c|}{$\text{AP}_\text{3D}$ IoU>0.25}  & \multicolumn{3}{c|}{$\text{AP}_\text{BEV}$ IoU>0.25} \\
      \cline{2-13}
      &Easy&Moderate&Hard&Easy&Moderate&Hard&Easy&Moderate&Hard&Easy&Moderate&Hard\\
      \hline
      Pedestrian & 10.33 & 10.12 & 9.03 & 11.77&11.30&11.30 &23.19&19.18&18.29&23.59&19.36&18.58\\
      Cyclist &16.01 & 11.44 & 11.26 & 16.87&11.67&11.58&32.05 & 19.97 & 18.48 & 32.59&20.09&18.69\\
      \hline
      \end{tabular}
    \label{tab:multi}
    \end{table}

\section{Prediction in Heatmaps}
\begin{figure}[htb]
  \begin{center}
    \includegraphics[width=0.8\columnwidth]{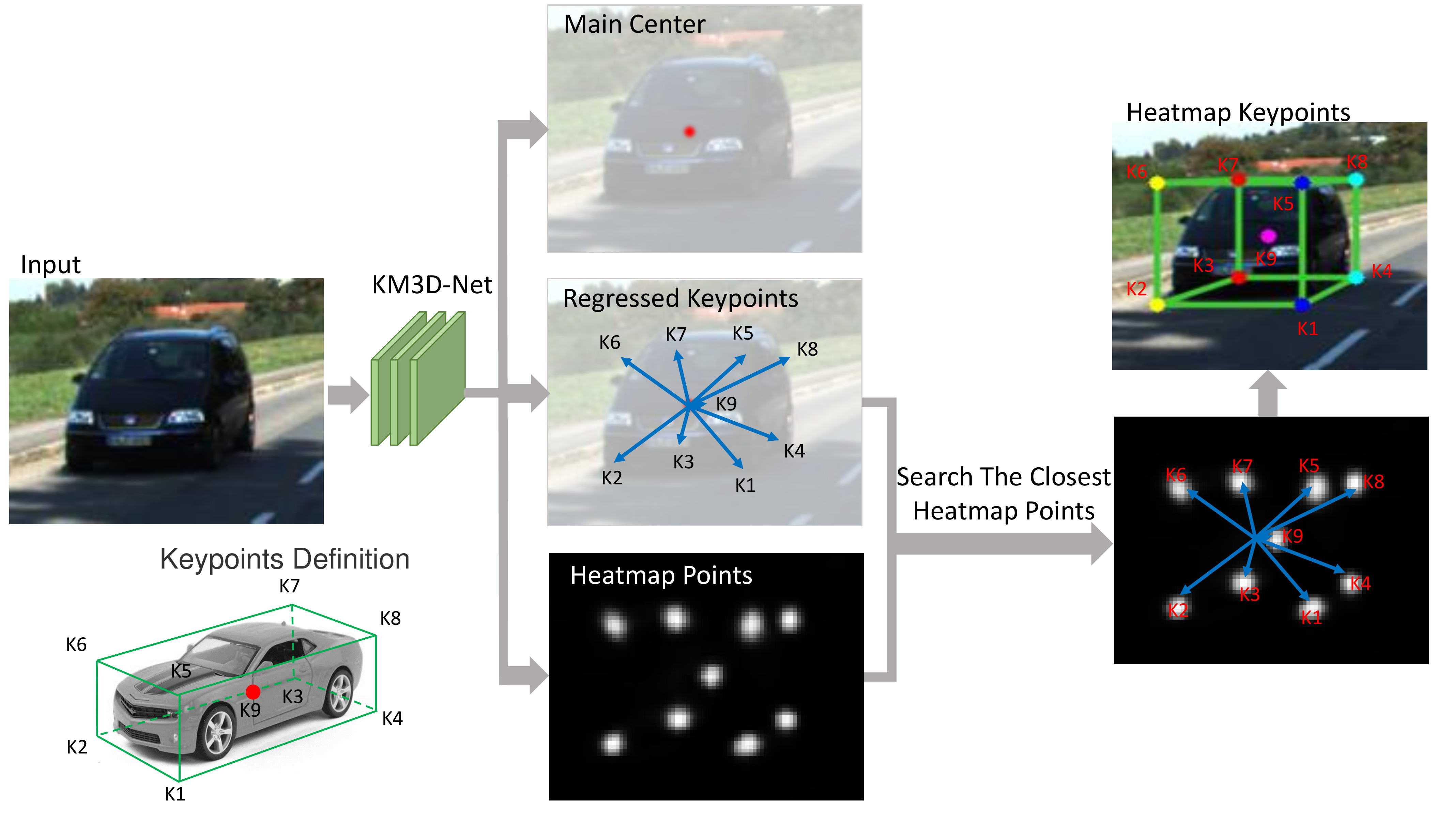}
  \end{center}
   \caption{Heatmap keypoints prediction. Here, our regressed coordinates of keypoints
   acts as a grouping cue, to assign individual heatmap points to their closest regressed keypoints.}
  \label{fig:heatmap}
\end{figure}
\begin{figure}[htb]
  \begin{center}
    \includegraphics[width=1\columnwidth]{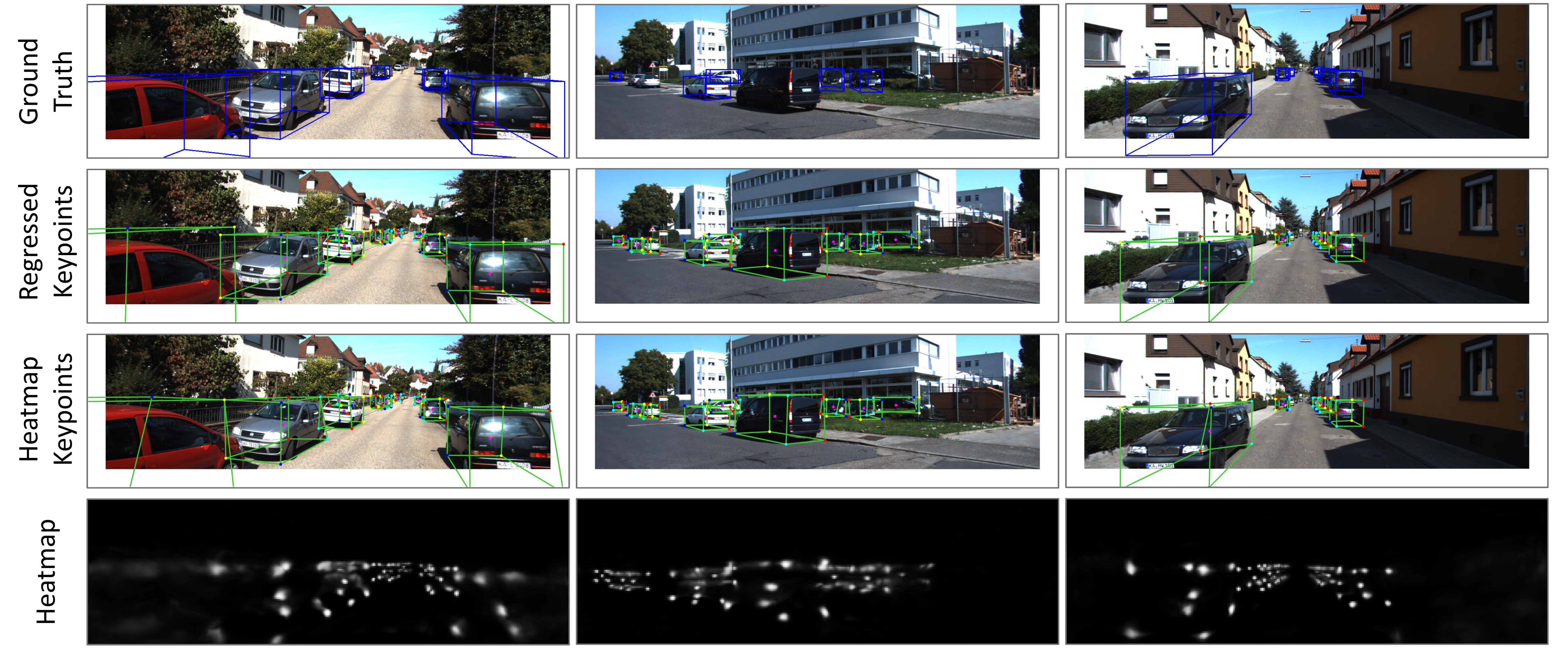}
  \end{center}
   \caption{Qualitative results. Regressed keypoints vs. heatmap keypoints.}
  \label{fig:heatmapvs}
\end{figure}

In this case, we add the heatmap $M_kh \in R^{\frac{h}{4} \times \frac{w}{4} \times 9}$ head parallel behind the backbone. Each channel corresponds to a kind of keypoint.
We train these heatmaps by focal loss. As shown in Fig. \ref{fig:heatmap}, In the inference phase, we extract heatmap points by imposing a $3 \times 3$ max-pooling following the main center filter. Then, we assign the regressed coordinates of keypoints to their closest heatmap points to group these in an object. Their qualitative results can be seen in Fig. \ref{fig:heatmapvs}. Note that heatmap have no ability to generate truncated keypoints, and we supplement these keypoints by regressed keypoints. Nevertheless, qualitative results show that the regressed keypoints have a better effect than heatmap keypoints. We believe that regressed keypoints are easy to establish relationships with each other. For example, the keypoints above the vehicle are often similar to the X-axis coordinates of the keypoints below.
\section{Failure Cases}
\begin{figure}[htb]
  \begin{center}
    \includegraphics[width=1\columnwidth]{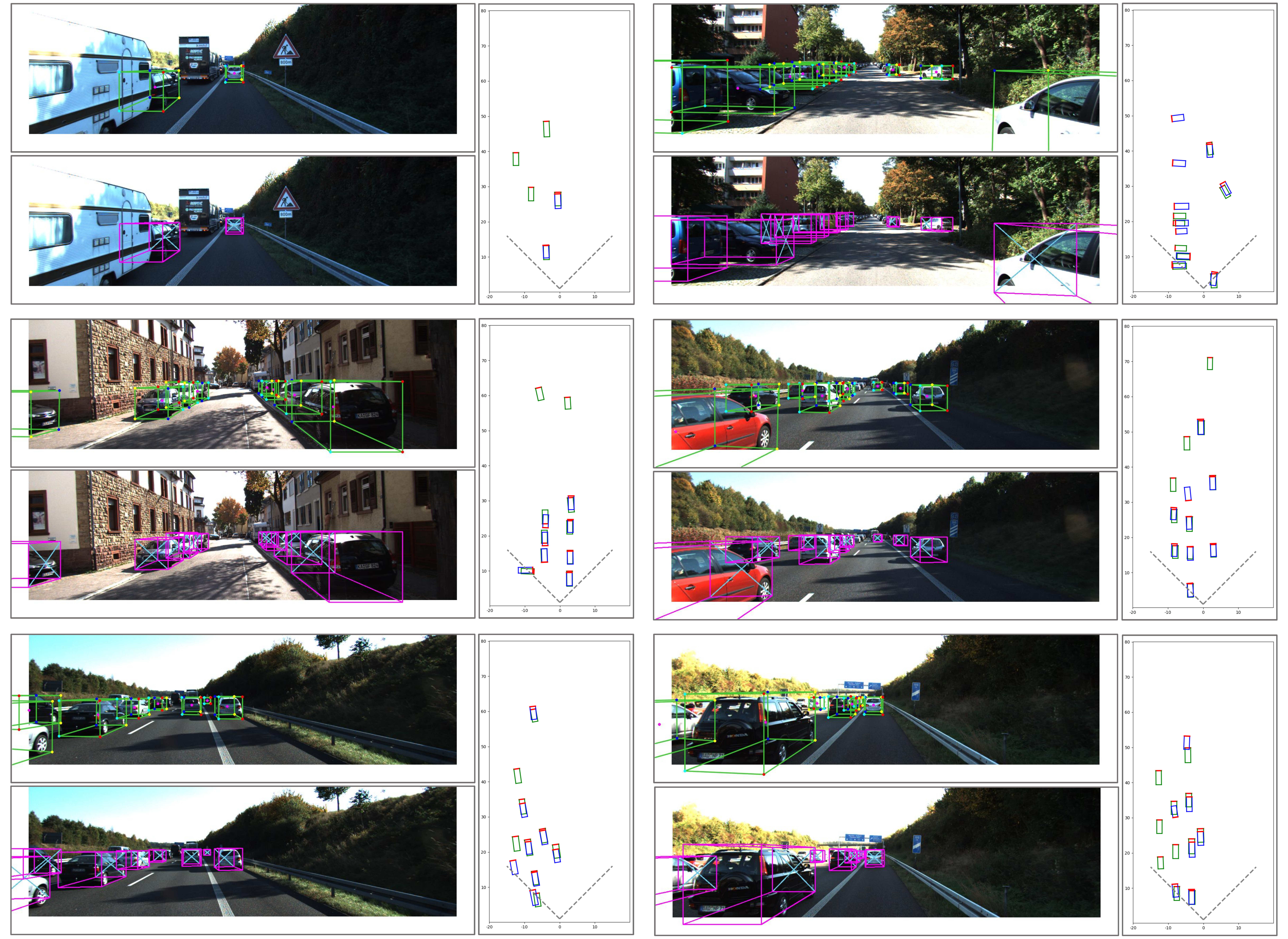}
  \end{center}
   \caption{Some failure cases. Most failures occur in very congested or distant situations. }
  \label{fig:failure}
\end{figure}

As shown in in Fig. \ref{fig:failure}, most failures occur in very congested or distant situations where objects dimension and orientation are difficult to predict based only on image information. Image-based methods have all these inherent limitations, which make their performance have a big gap to LiDAR-based methods. Nevertheless, we still believe that a better trade-off between speed and precision could make an image-based approach practical, as people can accurately and efficiently detect objects without relying on LiDAR system.

\clearpage
\bibliographystyle{splncs}
\bibliography{ref}

\begin{thebibliography}{10}

\bibitem{chen2017multi}
Chen, X., Ma, H., Wan, J., Li, B., Xia, T.:
\newblock Multi-view 3d object detection network for autonomous driving.
\newblock In: Proceedings of the IEEE Conference on Computer Vision and Pattern
  Recognition. (2017)  1907--1915

\bibitem{zhou2018voxelnet}
Zhou, Y., Tuzel, O.:
\newblock Voxelnet: End-to-end learning for point cloud based 3d object
  detection.
\newblock In: Proceedings of the IEEE Conference on Computer Vision and Pattern
  Recognition. (2018)  4490--4499

\bibitem{yang2018pixor}
Yang, B., Luo, W., Urtasun, R.:
\newblock Pixor: Real-time 3d object detection from point clouds.
\newblock In: Proceedings of the IEEE conference on Computer Vision and Pattern
  Recognition. (2018)  7652--7660

\bibitem{qi2018frustum}
Qi, C.R., Liu, W., Wu, C., Su, H., Guibas, L.J.:
\newblock Frustum pointnets for 3d object detection from rgb-d data.
\newblock In: Proceedings of the IEEE Conference on Computer Vision and Pattern
  Recognition. (2018)  918--927

\bibitem{shi2019pointrcnn}
Shi, S., Wang, X., Li, H.:
\newblock Pointrcnn: 3d object proposal generation and detection from point
  cloud.
\newblock In: Proceedings of the IEEE Conference on Computer Vision and Pattern
  Recognition. (2019)  770--779

\bibitem{brazil2019m3drpn}
Brazil, G., Liu, X.:
\newblock M3d-rpn: Monocular 3d region proposal network for object detection.
\newblock In: Proceedings of the IEEE International Conference on Computer
  Vision, Seoul, South Korea (2019)

\bibitem{chen2016monocular}
Chen, X., Kundu, K., Zhang, Z., Ma, H., Fidler, S., Urtasun, R.:
\newblock Monocular 3d object detection for autonomous driving.
\newblock In: Proceedings of the IEEE Conference on Computer Vision and Pattern
  Recognition. (2016)  2147--2156

\bibitem{manhardt2019roi}
Manhardt, F., Kehl, W., Gaidon, A.:
\newblock Roi-10d: Monocular lifting of 2d detection to 6d pose and metric
  shape.
\newblock In: Proceedings of the IEEE Conference on Computer Vision and Pattern
  Recognition. (2019)  2069--2078

\bibitem{mousavian20173d}
Mousavian, A., Anguelov, D., Flynn, J., Kosecka, J.:
\newblock 3d bounding box estimation using deep learning and geometry.
\newblock In: Proceedings of the IEEE Conference on Computer Vision and Pattern
  Recognition. (2017)  7074--7082

\bibitem{qin2019monogrnet}
Qin, Z., Wang, J., Lu, Y.:
\newblock Monogrnet: A geometric reasoning network for monocular 3d object
  localization.
\newblock In: Proceedings of the AAAI Conference on Artificial Intelligence.
  Volume~33. (2019)  8851--8858

\bibitem{simonelli2019disentangling}
Simonelli, A., Bulo, S.R., Porzi, L., L{\'o}pez-Antequera, M., Kontschieder,
  P.:
\newblock Disentangling monocular 3d object detection.
\newblock In: Proceedings of the IEEE International Conference on Computer
  Vision. (2019)  1991--1999

\bibitem{li2020rtm3d}
Li, P., Zhao, H., Liu, P., Cao, F.:
\newblock Rtm3d: Real-time monocular 3d detection from object keypoints for
  autonomous driving (2020)

\bibitem{he2019mono3d++}
He, T., Soatto, S.:
\newblock Mono3d++: Monocular 3d vehicle detection with two-scale 3d hypotheses
  and task priors.
\newblock In: Proceedings of the AAAI Conference on Artificial Intelligence.
  Volume~33. (2019)  8409--8416

\bibitem{xu2018multi}
Xu, B., Chen, Z.:
\newblock Multi-level fusion based 3d object detection from monocular images.
\newblock In: Proceedings of the IEEE Conference on Computer Vision and Pattern
  Recognition. (2018)  2345--2353

\bibitem{Li_2019_CVPR}
Li, B., Ouyang, W., Sheng, L., Zeng, X., Wang, X.:
\newblock Gs3d: An efficient 3d object detection framework for autonomous
  driving.
\newblock In: The IEEE Conference on Computer Vision and Pattern Recognition
  (CVPR). (June 2019)

\bibitem{zhou2019objects}
Zhou, X., Wang, D., Kr{\"a}henb{\"u}hl, P.:
\newblock Objects as points.
\newblock In: arXiv preprint arXiv:1904.07850. (2019)

\bibitem{Naiden2019ShiftRD}
Naiden, A., Paunescu, V., Kim, G., Jeon, B., Leordeanu, M.:
\newblock Shift r-cnn: Deep monocular 3d object detection with closed-form
  geometric constraints.
\newblock 2019 IEEE International Conference on Image Processing (ICIP) (2019)
  61--65

\bibitem{liu2019deep}
Liu, L., Lu, J., Xu, C., Tian, Q., Zhou, J.:
\newblock Deep fitting degree scoring network for monocular 3d object
  detection.
\newblock In: Proceedings of the IEEE Conference on Computer Vision and Pattern
  Recognition. (2019)  1057--1066

\bibitem{li2019stereo}
Li, P., Chen, X., Shen, S.:
\newblock Stereo r-cnn based 3d object detection for autonomous driving.
\newblock In: Proceedings of the IEEE Conference on Computer Vision and Pattern
  Recognition. (2019)  7644--7652

\bibitem{Berthelot2019MixMatch}
Berthelot, D., Carlini, N., Goodfellow, I., Papernot, N., Oliver, A., Raffel,
  C.A.:
\newblock Mixmatch: A holistic approach to semi-supervised learning.
\newblock In: NeurIPS. (2019)

\bibitem{laine2016temporal}
Laine, S., Aila, T.:
\newblock Temporal ensembling for semi-supervised learning.
\newblock In: ICLR. (2017)

\bibitem{NIPS2015_5947}
Rasmus, A., Berglund, M., Honkala, M., Valpola, H., Raiko, T.:
\newblock Semi-supervised learning with ladder networks.
\newblock In Cortes, C., Lawrence, N.D., Lee, D.D., Sugiyama, M., Garnett, R.,
  eds.: Advances in Neural Information Processing Systems 28.
\newblock Curran Associates, Inc. (2015)  3546--3554

\bibitem{tarvainen2017mean}
Tarvainen, A., Valpola, H.:
\newblock Mean teachers are better role models: Weight-averaged consistency
  targets improve semi-supervised deep learning results.
\newblock In: NeurIPS. (2017)

\bibitem{srivastava2014dropout}
Srivastava, N., Hinton, G.E., Krizhevsky, A., Sutskever, I., Salakhutdinov, R.:
\newblock Dropout: a simple way to prevent neural networks from overfitting.
\newblock Journal of Machine Learning Research \textbf{15}(1) (2014)
  1929--1958

\bibitem{ma2019accurate}
Ma, X., Wang, Z., Li, H., Zhang, P., Ouyang, W., Fan, X.:
\newblock Accurate monocular 3d object detection via color-embedded 3d
  reconstruction for autonomous driving.
\newblock In: Proceedings of the IEEE International Conference on Computer
  Vision. (2019)  6851--6860

\bibitem{wang2019pseudo}
Wang, Y., Chao, W.L., Garg, D., Hariharan, B., Campbell, M., Weinberger, K.Q.:
\newblock Pseudo-lidar from visual depth estimation: Bridging the gap in 3d
  object detection for autonomous driving.
\newblock In: Proceedings of the IEEE Conference on Computer Vision and Pattern
  Recognition. (2019)  8445--8453

\bibitem{chabot2017deep}
Chabot, F., Chaouch, M., Rabarisoa, J., Teuli{\`e}re, C., Chateau, T.:
\newblock Deep manta: A coarse-to-fine many-task network for joint 2d and 3d
  vehicle analysis from monocular image.
\newblock In: Proceedings of the IEEE Conference on Computer Vision and Pattern
  Recognition. (2017)  2040--2049

\bibitem{zeeshan2014cars}
Zeeshan~Zia, M., Stark, M., Schindler, K.:
\newblock Are cars just 3d boxes?-jointly estimating the 3d shape of multiple
  objects.
\newblock In: Proceedings of the IEEE Conference on Computer Vision and Pattern
  Recognition. (2014)  3678--3685

\bibitem{murthy2017reconstructing}
Murthy, J.K., Krishna, G.S., Chhaya, F., Krishna, K.M.:
\newblock Reconstructing vehicles from a single image: Shape priors for road
  scene understanding.
\newblock In: 2017 IEEE International Conference on Robotics and Automation
  (ICRA), IEEE (2017)  724--731

\bibitem{ren2015faster}
Ren, S., He, K., Girshick, R., Sun, J.:
\newblock Faster r-cnn: Towards real-time object detection with region proposal
  networks.
\newblock In: Advances in neural information processing systems. (2015)  91--99

\bibitem{reddy2018semi-supervised}
Reddy, Y.C.A.P., Viswanath, P., Reddy, B.E.:
\newblock Semi-supervised learning: a brief review.
\newblock International journal of engineering and technology \textbf{7} (2018)
  ~81

\bibitem{tu2019a}
Tu, E., Yang, J.:
\newblock A review of semi supervised learning theories and recent advances.
\newblock arXiv: Learning (2019)

\bibitem{lee2013pseudo}
Lee, D.H.:
\newblock Pseudo-label: The simple and efficient semi-supervised learning
  method for deep neural networks.
\newblock In: ICML Workshops. (2013)

\bibitem{rasmus2015semi}
Rasmus, A., Berglund, M., Honkala, M., Valpola, H., Raiko, T.:
\newblock Semi-supervised learning with ladder networks.
\newblock In: NeurIPS. (2015)

\bibitem{sajjadi2016mutual}
Sajjadi, M., Javanmardi, M., Tasdizen, T.:
\newblock Mutual exclusivity loss for semi-supervised deep learning.
\newblock In: ICIP. (2016)

\bibitem{miyato2018virtual}
Miyato, T., Maeda, S.i., Ishii, S., Koyama, M.:
\newblock Virtual adversarial training: a regularization method for supervised
  and semi-supervised learning.
\newblock T-PAMI (2018)

\bibitem{xie2019unsupervised}
Xie, Q., Dai, Z., Hovy, E., Luong, M.T., Le, Q.V.:
\newblock Unsupervised data augmentation for consistency training.
\newblock arXiv preprint arXiv:1904.12848 (2019)

\bibitem{xie2019self}
Xie, Q., Hovy, E., Luong, M.T., Le, Q.V.:
\newblock Self-training with noisy student improves imagenet classification.
\newblock arXiv preprint arXiv:1911.04252 (2019)

\bibitem{sohn2020fixmatch}
Sohn, K., Berthelot, D., Li, C.L., Zhang, Z., Carlini, N., Cubuk, E.D.,
  Kurakin, A., Zhang, H., Raffel, C.:
\newblock Fixmatch: Simplifying semi-supervised learning with consistency and
  confidence.
\newblock arXiv preprint arXiv:2001.07685 (2020)

\bibitem{Lecun98gradient-basedlearning}
Lecun, Y., Bottou, L., Bengio, Y., Haffner, P.:
\newblock Gradient-based learning applied to document recognition.
\newblock In: Proceedings of the IEEE. (1998)  2278--2324

\bibitem{law2018cornernet}
Law, H., Deng, J.:
\newblock Cornernet: Detecting objects as paired keypoints.
\newblock In: Proceedings of the European Conference on Computer Vision (ECCV).
  (2018)  734--750

\bibitem{Giles2008An}
Giles, M.B.:
\newblock An extended collection of matrix derivative results for forward and
  reverse mode automatic differentiation.
\newblock (2008)

\bibitem{Ionescu2015Matrix}
Ionescu, C., Vantzos, O., Sminchisescu, C.:
\newblock Matrix backpropagation for deep networks with structured layers.
\newblock In: IEEE International Conference on Computer Vision. (2015)

\bibitem{geiger2012we}
Geiger, A., Lenz, P., Urtasun, R.:
\newblock Are we ready for autonomous driving? the kitti vision benchmark
  suite.
\newblock In: 2012 IEEE Conference on Computer Vision and Pattern Recognition,
  IEEE (2012)  3354--3361

\bibitem{he2016deep}
He, K., Zhang, X., Ren, S., Sun, J.:
\newblock Deep residual learning for image recognition.
\newblock In: Proceedings of the IEEE conference on computer vision and pattern
  recognition. (2016)  770--778

\bibitem{yu2018deep}
Yu, F., Wang, D., Shelhamer, E., Darrell, T.:
\newblock Deep layer aggregation.
\newblock In: Proceedings of the IEEE Conference on Computer Vision and Pattern
  Recognition. (2018)  2403--2412

\bibitem{chen20173d}
Chen, X., Kundu, K., Zhu, Y., Ma, H., Fidler, S., Urtasun, R.:
\newblock 3d object proposals using stereo imagery for accurate object class
  detection.
\newblock IEEE transactions on pattern analysis and machine intelligence
  \textbf{40}(5) (2017)  1259--1272

\bibitem{cai2020monocular}
Cai, Y., Li, B., Jiao, Z., Li, H., Zeng, X., Wang, X.:
\newblock Monocular 3d object detection with decoupled structured polygon
  estimation and height-guided depth estimation.
\newblock AAAI (2020)

\bibitem{qin2019tlnet}
Qin, Z., Wang, J., Lu, Y.:
\newblock Triangulation learning network: from monocular to stereo 3d object
  detection.
\newblock IEEE Conference on Computer Vision and Pattern Recognition (CVPR)
  (2019)

\bibitem{monofenet}
Bao, W., Xu, B., Chen, Z.:
\newblock Monofenet: Monocular 3d object detection with feature enhancement
  networks.
\newblock IEEE Transactions on Image Processing (2019)

\bibitem{ku2019monopsr}
Ku*, J., Pon*, A.D., Waslander, S.L.:
\newblock Monocular 3d object detection leveraging accurate proposals and shape
  reconstruction.
\newblock In: CVPR. (2019)

\bibitem{licvpr2019}
Li, P., Chen, X., Shen, S.:
\newblock Stereo r-cnn based 3d object detection for autonomous driving.
\newblock In: CVPR. (2019)

\end{thebibliography}
\end{document}